\documentclass[conference]{IEEEtran}

\usepackage[T1]{fontenc}

\usepackage{amsmath,amssymb,amsfonts}
\usepackage{pifont}

\usepackage{graphicx}
\usepackage{xcolor}
\usepackage{tabularx}
\usepackage{multirow}
\usepackage{array}

\usepackage{algorithmic}
\usepackage{listings}

\usepackage{cite}

\usepackage{siunitx}
\usepackage[hidelinks]{hyperref}
\usepackage{orcidlink}

\def\BibTeX{{\rm B\kern-.05em{\sc i\kern-.025em b}\kern-.08em
    T\kern-.1667em\lower.7ex\hbox{E}\kern-.125emX}
}

\begin{document}

\bstctlcite{control}

\title{CienaLLM: Generative Climate-Impact Extraction from News Articles with Autoregressive LLMs}

\author{
    \IEEEauthorblockN{Javier Vela-Tambo\IEEEauthorrefmark{1}\IEEEauthorrefmark{3}, Jorge Gracia\IEEEauthorrefmark{2}, and Fernando Dominguez-Castro\IEEEauthorrefmark{3}}

    \vspace{0.05in}

    \IEEEauthorblockA{\IEEEauthorrefmark{1}Worcester Polytechnic Institute, Worcester, MA, USA. \emph{jvela@wpi.edu} \orcidlink{0000-0002-6818-9191}}

    \IEEEauthorblockA{\IEEEauthorrefmark{2}Aragon Institute of Engineering Research, Universidad de Zaragoza, Spain. \emph{jogracia@unizar.es} \orcidlink{0000-0001-6452-7627}}

    \IEEEauthorblockA{\IEEEauthorrefmark{3}Pyrenean Institute of Ecology (IPE-CSIC), Zaragoza, Spain. \emph{fdominguez@ipe.csic.es} \orcidlink{0000-0003-3085-7040}}
}

\maketitle

\thispagestyle{plain}
\pagestyle{plain}

\begin{abstract}
    Understanding and monitoring the socio-economic impacts of climate hazards requires extracting structured information from heterogeneous news articles on a large scale. To that end, we have developed CienaLLM, a modular framework based on schema-guided Generative Information Extraction. CienaLLM uses open-weight Large Language Models for zero-shot information extraction from news articles, and supports configurable prompts and output schemas, multi-step pipelines, and cloud or on-premise inference. To systematically assess how the choice of LLM family, size, precision regime, and prompting strategy affect performance, we run a large factorial study in models, precisions, and prompt engineering techniques. An additional response parsing step nearly eliminates format errors while preserving accuracy; larger models deliver the strongest and most stable performance, while quantization offers substantial efficiency gains with modest accuracy trade-offs; and prompt strategies show heterogeneous, model-specific effects. CienaLLM matches or outperforms the supervised baseline in accuracy for extracting drought impacts from Spanish news, although at a higher inference cost. While evaluated in droughts, the schema-driven and model-agnostic design is suitable for adapting to related information extraction tasks (e.g., other hazards, sectors, or languages) by editing prompts and schemas rather than retraining. We release code, configurations, and schemas to support reproducible use.
\end{abstract}


\section{Introduction and Related Work} \label{sec:intro}

Extreme climate and weather events, including floods, hailstorms, and heatwaves, are among the most disruptive manifestations of climate change, producing severe socio-economic and environmental consequences. Such events can damage ecosystems, threaten food security, and in the most severe cases endanger lives~\cite{worldmeteorologicalorganizationwmoWMOAtlasMortality2021}. Among these hazards, drought is responsible for one of the largest losses, reducing agricultural production, stressing the water and energy systems, and increasing social vulnerability~\cite{pena-gallardoImpactDroughtProductivity2019, vanvlietImpactsRecentDrought2016, zhaoResponsesHydroelectricityGeneration2023, intergovernmentalpanelonclimatechangeipccClimateChange20222023}. With climate change increasing both the frequency and severity of drought events~\cite{vicente-serranoGlobalDroughtTrends2022,vicente-serranoUnravelingInfluenceAtmospheric2020}, systematic knowledge of their evolution is urgently needed.

Traditional drought indices, such as the Standardized Precipitation Index or the Palmer Drought Severity Index~\cite{svoboda2016handbook,hess-20-2589-2016,wilhiteChapter1Drought2000}, capture the physical evolution of drought events, but do not reflect their social and ecological impacts. Understanding these dimensions requires information on the impacts that droughts produce across sectors and regions. Drought impacts databases like the European Drought Impact Report Inventory~\cite{stahlImpactsEuropeanDrought2016} provide valuable evidence but remain uneven in coverage and detail, limiting systematic monitoring and adaptation planning. By contrast, newspapers have documented the consequences of droughts for decades at national or regional scale, making them one of the richest sources of impact information~\cite{boykoffClimateChangeJournalistic2007}. However, the impact information in newspapers is highly dispersed and heterogeneous. Previous studies have used manual or semi-automatic content analysis of news articles to reconstruct drought events and their impacts, including those carried out in Ireland~\cite{oconnorRelatingDroughtIndices2023,evaIrishDroughtImpacts2022}, the United States~\cite{dowNewsCoverageDrought2010}, Australia~\cite{bellDriestContinentGreediest2009,hurlimannNewspaperCoverageWater2012,rutledge-priorDroughtsFleetingRains2021}, the United Kingdom~\cite{dayrellRepresentationDroughtEvents2022}, and Spain~\cite{llasatAnalysisEvolutionHydrometeorological2009,ruizsinogaDroughtsTheirSocial2013}. These efforts underscore the potential of news as an impact source, but their reliance on human annotation limits both scale and generalizability, restricting analyses to narrow regions or short timeframes.

Automatically extracting structured information from news articles is a challenging task. They are often long and heterogeneous, combining relevant details with unrelated narrative elements. Impacts are described in varied and nuanced ways, and geographic references are often implicit, appearing through mentions of rivers, reservoirs, or towns rather than administrative units, and therefore require contextual inference~\cite{leidnerToponymResolutionText2007,delozierCreatingNovelGeolocation2016,grunewaldWorkingNamedPlaces2024}. These challenges are not limited to drought: similar difficulties arise when analyzing other events such as floods, hailstorms, or heatwaves, or when seeking related information.

Transformer-based models such as BERT and its derivatives leverage self-attention to capture context, enabling fine-tuned supervised systems that learn rich representations from text and achieve strong task performance~\cite{vaswaniAttentionAllYou2023,devlinBERTPretrainingDeep2019,liuRoBERTaRobustlyOptimized2019}. This has enabled the training of supervised models that build on autoencoding language models, allowing them to learn rich contextual representations and achieve strong task performance. In the climate domain, supervised NLP approaches have been widely applied, covering tasks such as classifying article relevance, assigning impact categories, and recognizing toponyms and sectoral entities. Many of these systems build on machine learning methods, including encoder-based language models with task-specific heads trained on carefully curated datasets~\cite{sodogeAutomatizedSpatiotemporalDetection2023,pitacostaImprovedKnowledgeWaterrelated2024,duarteApplicationNaturalLanguage2024,zhangTweetDroughtDeepLearningDrought2022,domalaAutomatedIdentificationDisaster2020,zouMulticlassMultilabelClassification2024,laiNaturalLanguageProcessing2022,otudiClassifyingSevereWeather2023}.

López-Otal et al.~\cite{lopez-otalSeqIAPythonFramework2025} introduce SeqIA, a supervised approach that trains separate classifiers on top of Transformer encoders to detect drought-relevant articles and to classify their impacts in Spanish news articles. SeqIA achieves strong in-domain performance, but, like other supervised systems, it depends on new annotations and retraining when applied to different hazards, languages, regions, or information schemas. Moreover, although the system attempts location extraction, it identifies all locations mentioned rather than isolating only those directly affected by drought, a distinction that is crucial for spatial analysis. Accurately extracting the truly impacted locations is challenging, as it often requires inferring them from indirect references such as rivers, reservoirs, or nearby towns.

Autoregressive generative Large Language Models (LLMs), such as GTP-4~\cite{openaiGPT4TechnicalReport2024}, or Llama~\cite{touvronLLaMAOpenEfficient2023a}, represent a shift beyond encoder-based approaches. Rather than producing fixed representations for downstream classifiers, these models are trained on massive text corpora to generate text token by token, capturing broad contextual and world knowledge. Unlike supervised encoders, which require task-specific retraining, generative LLMs can adapt to new tasks with little or no additional supervision, making them highly versatile across domains~\cite{brownLanguageModelsAre2020,bubeckSparksArtificialGeneral2023}.

Schema-guided Generative Information Extraction (GenIE) builds directly on this capability by prompting LLMs to output structured information aligned to a predefined schema (e.g., impact type, location) in JSON format~\cite{xuLargeLanguageModels2024a,baiSchemaDrivenInformationExtraction2023,wangSLOTStructuringOutput2025,josifoskiGenIEGenerativeInformation2022b,rettenbergerUsingLargeLanguage2024,weiChatIEZeroShotInformation2024a,shiriDecomposeEnrichExtract2024,popovicDocIEXLLM25InContextLearning2025}. This paradigm eliminates the need for task-specific fine-tuning and enables rapid adaptation to evolving schemas and cross-domain applications. GenIE has shown promise in areas as diverse as health, biomedicine, and chemistry~\cite{dagdelenStructuredInformationExtraction2024,liuImprovingLLMBasedHealth2024,wiestLLMAIxOpenSource2024,hsuLLMIEPythonPackage2025}. Li et al.~\cite{liUsingLLMsBuild2024} have explored this paradigm for climate impacts, demonstrating feasibility. However, their work uses a fixed taxonomy (EM-DAT~\cite{delforgeEMDATEmergencyEvents2023}), evaluates only a narrow LLM coverage, and relies on Wikipedia, a more structured and unambiguous source of information.

We adopt a schema-guided GenIE approach to process drought-related news from Spain, addressing three tasks: (i) detecting article relevance, (ii) identifying multiple impact types, and (iii) extracting impacted locations. We use open-weight LLMs, whose weights are downloadable for local inference (e.g. Gemma~\cite{teamGemmaOpenModels2024}, Llama~\cite{touvronLLaMAOpenEfficient2023a}, Qwen~\cite{qwenQwen25TechnicalReport2025}). This enables on-premise experiments through ecosystems such as Ollama\footnote{\url{https://ollama.com/}}, improving privacy, controllability, reproducibility, and often cost and energy efficiency in low-resource settings. These advantages stand in contrast to closed API-only systems (e.g., GPT-4~\cite{openaiGPT4TechnicalReport2024}, Gemini~\cite{teamGeminiFamilyHighly2025}). We compare model families and model sizes, using the number of parameters as a proxy for capacity. We also compare full-precision against quantized precision regimes. Quantization reduces latency and memory requirements while introducing only limited accuracy loss~\cite{zhuSurveyModelCompression2024,dettmersQLoRAEfficientFinetuning2023}. In addition, we analyze the effect of different prompt engineering strategies~\cite{schulhoffPromptReportSystematic2024}. Our evaluation focuses on three dimensions: accuracy (F1, precision, recall), efficiency (latency and compute), and reliability. Ensuring response reliability is challenging, as models may produce malformed JSON~\cite{luLearningGenerateStructured2025}, hallucinate information~\cite{linTruthfulQAMeasuringHow2022,jiSurveyHallucinationNatural2023}, or show sensitivity to prompt wording~\cite{ribeiroAccuracyBehavioralTesting2020,zhuPromptBenchUnifiedLibrary2024}.

Despite promising demonstrations of GenIE~\cite{liUsingLLMsBuild2024}, no systematic evaluation has examined the capabilities of open-weight LLMs for climate-impact extraction from news. Key open questions remain: (i) how model family, size, and precision regime influence accuracy, efficiency, and reliability; (ii) how consistently models generate parsable, schema-conformant outputs; (iii) whether prompt engineering strategies can meaningfully improve extraction; and (iv) what performance–efficiency trade-offs arise across different models and configurations.

In this work, we pursue GenIE in a zero-shot setting: rather than training new classifiers, we rely on the general pre-trained knowledge of LLMs and guide them solely with natural language prompts to extract structured information. A key advantage of this design is that the system inherits improvements in new LLM releases, including reasoning ability, factual coverage, and structured output reliability. These benefits are obtained without requiring additional annotation or retraining. We evaluate this approach on drought as a representative case while designing methods intended to generalize to other climate hazards and domains.

To address these gaps, we contribute the following.

\begin{enumerate}
    \item \textbf{Framework}. We introduce CienaLLM\footnote{\url{https://github.com/lcsc/ciena_llm}} (Climate Impact Extraction from News Articles using LLMs), a modular, open-source toolkit for schema-guided GenIE.
    \item \textbf{Evaluation}. We present a large-scale factorial study covering 384 configurations across 12 open-weight models, spanning families, sizes, and quantization regimes, under controlled prompting and parsing strategies.
    \item \textbf{Comparison}. We benchmark CienaLLM against SeqIA~\cite{lopez-otalSeqIAPythonFramework2025} on shared datasets for drought impact extraction from Spanish news articles.
    \item \textbf{Insights}. We provide a detailed analysis of performance–efficiency trade-offs, parsing reliability, and quantization viability, identifying which prompt strategies are most effective for different model families and sizes.
    \item \textbf{Release}. We release code, configurations, and schema definitions to support reproducible research.
\end{enumerate}

The remainder of this paper is organized as follows: Section~\ref{sec:datasets} introduces the datasets, Section~\ref{sec:methods} presents the methodology and the CienaLLM framework, Section~\ref{sec:experiments} describes the experimental setup, and Section~\ref{sec:results} reports the results. Finally, Section~\ref{sec:discussion} discusses the findings and Section~\ref{sec:conclusions} concludes with directions for future work.

\newpage

\section{Datasets} \label{sec:datasets}

This study builds on and extends the datasets compiled by López-Otal et al.~\cite{lopez-otalSeqIAPythonFramework2025}. Their collection~\cite{lopezotalSeqIAAnnotatedDroughtrelated2025} includes two components: entire news articles labeled for drought relevance, and individual sentences from other articles annotated for specific drought impacts. We reuse these datasets but add new annotations that enable article-level impact extraction, spatial localization of drought effects, and mitigation of impacts imbalance.

To complement these datasets, we have downloaded the entire online archives of major Spanish outlets to assemble broader news corpora (see Appendix~\ref{sec:appendix_news}). Combining national and regional sources ensures coverage across scales, making the corpora well suited for future large-scale analyses of climate-related impacts in Spain using the extraction tools developed in this study. All articles follow a standardized structure based on the \textit{NewsArticle} schema\footnote{\url{https://schema.org/NewsArticle}}, which provides consistent metadata such as headline, body text, publication date, and URL.

\subsection{Drought Impacts Dataset} \label{sec:datasets_did}

To evaluate the performance of our system in extracting drought-related impacts from news articles, we use two dataset collections: the \textit{Drought Impact Identification Training Datasets} and the \textit{End-to-End} (E2E) dataset~\cite{lopezotalSeqIAAnnotatedDroughtrelated2025}. Both have been reannotated and adapted in our work to support article-level impact evaluation. The news articles were sourced from \textit{El País}\footnote{\url{https://elpais.com}} and Grupo Z, now acquired by Prensa Ibérica\footnote{\url{https://www.prensaiberica.es}}.

For the \textit{Drought Impact Identification Training Datasets}, we reannotated a subset of 244 articles at the article level, assigning one or more impact labels based on the presence of relevant information anywhere in the article, rather than by sentence. Drought relevance annotations and articles from underepresented impact types were added to the original dataset. An important caveat of the original datasets is that the same article could contribute a sentence to the training split for one impact and to the test split for another, which limits direct comparability with SeqIA. The E2E dataset was also reannotated at the article level following the same criteria. The substantial inter-annotator agreement (Cohen’s kappa = $0.695 \pm 0.140$) with a second expert who reannotated the E2E dataset at the article level highlights the inherent difficulty of annotating drought impacts in news articles~\cite{landisMeasurementObserverAgreement1977}.

Finally, we merged both reannotated datasets into a single evaluation resource, the \textit{Drought Impacts Dataset} (DID). Only articles labeled as drought-related were retained, as the objective is to assess impact extraction after relevance filtering. The final dataset comprises 386 annotated articles with multi-label assignments for the four impact types: agriculture, livestock, hydrological resources, and energy. We applied a stratified 70/30 split into validation and test subsets, excluding combinations of labels with fewer than two samples to ensure meaningful stratification. Since this dataset is only used for evaluation, no training split was defined. Table~\ref{tab:datasets_did} summarizes the composition of the merged dataset and its split.

\begin{table}[htbp]
    \caption{Distribution of class labels across splits in the Drought Impacts Dataset.}
    \label{tab:datasets_did}
    \centering
    \begin{tabular}{|l|r|r|r|}
        \hline
        \textbf{Label}          & \textbf{Validation} & \textbf{Test} & \textbf{Total} \\
        \hline
        Agriculture             & 126                 & 55            & 181            \\
        Livestock               & 67                  & 30            & 97             \\
        Hydrological Resources  & 128                 & 56            & 184            \\
        Energy                  & 42                  & 17            & 59             \\
        \hline
        \textbf{Total Articles} & 269                 & 117           & 386            \\
        \hline
    \end{tabular}
\end{table}

\subsection{Drought Relevance Dataset} \label{sec:datasets_drd}

The \textit{Drought Relevance Dataset} (DRD)~\cite{lopez-otalSeqIAPythonFramework2025} consists of 2,240 news articles from El País, published between 1976 and 2023, and labeled as either drought-related (1,270) or not (970). The dataset was originally split into training and test subsets for supervised classification, as shown in Table~\ref{tab:datasets_drd}. We reuse this dataset to compare the performance of our LLM-based approach with that of a traditional supervised method. Both approaches aim to identify drought-relevant news articles, but differ in how relevance is determined: via a trained classifier or a prompted language model.

\begin{table}[htbp]
    \caption{Distribution of positive and negative labels across splits in the Drought Relevance Dataset.}
    \label{tab:datasets_drd}
    \centering
    \begin{tabular}{|l|r|r|r|}
        \hline
        \textbf{Label}          & \textbf{Train} & \textbf{Test} & \textbf{Total} \\
        \hline
        Positive                & 903            & 367           & 1270           \\
        Negative                & 665            & 305           & 970            \\
        \hline
        \textbf{Total Articles} & 1568           & 672           & 2240           \\
        \hline
    \end{tabular}
\end{table}

\subsection{Drought Impact Locations Dataset} \label{sec:datasets_dild}

To evaluate the ability of our system to extract the geographical scope of drought-related impacts, we created the \textit{Drought Impact Locations Dataset} (DILD). The DILD dataset consists of 100 drought-related articles from El País, annotated with the Spanish provinces affected by droughts as reported in news articles. We chose Spanish provinces as the target unit of analysis because drought impacts typically affect large regions rather than isolated municipalities, making provinces a suitable level of granularity for spatial annotation and evaluation, and is often used in drought research~\cite{pena-gallardoImpactDroughtProductivity2019,salvadorQuantificationEffectsDroughts2020}. Annotation was performed by a domain expert who assigned to each article the list of provinces where drought-related impacts were reported. In total, the dataset contains 835 annotated province-level entries across 100 articles, with each article linked to an average of 8.35 provinces. The number of provinces per article varies significantly, with a median of 1, but a maximum of 50 in some broad, national-level reports. All 50 Spanish provinces are represented in the dataset, ensuring full geographical coverage.

\section{Methodology and Framework} \label{sec:methods}

\subsection{Large Language Models} \label{sec:methods_llm}

We selected twelve open-weight LLMs covering three major model families: Gemma by Google\footnote{\url{https://deepmind.google/models/gemma/}}, Llama by Meta\footnote{\url{https://www.llama.com/}}, and Qwen by Alibaba\footnote{\url{https://chat.qwen.ai/}}. The selection spans three size tiers: small ($<7$B parameters), medium ($7-25$B), and large ($>25$B). It also includes two precision regimes: full-precision (unquantized; \texttt{fp16}) and 4-bit mixed-precision quantization (\texttt{q4\_K\_M}). To directly assess the effect of quantization, we include for each family an unquantized medium-sized model with the same configuration otherwise. All models were obtained from the Ollama model library\footnote{\url{https://ollama.com/library}} at the time of study. Table~\ref{tab:methods_llm} lists the main features of the selected models. In the table, and throughout the rest of this paper, each model configuration is given a unique identifier (\texttt{LLM}) that combines family, size, and quantization.

\begin{table*}[!t]
    \caption{Overview of the open-weight LLMs evaluated in this study.}
    \label{tab:methods_llm}
    \centering
    \begin{tabular}{|l|l|l|r|c|l|}
        \hline
        \textbf{\texttt{LLM}}  & \textbf{Model Name}            & \textbf{Family} & \textbf{Parameters (B) (Size)} & \textbf{Quantization} & \textbf{Release Date} \\
        \hline
        \texttt{gemma\_4b}     & gemma3:4b-it-q4\_K\_M          & Gemma           & 4 (S)                          & \texttt{q4\_K\_M}     & March 10, 2025        \\
        \texttt{gemma\_12b}    & gemma3:12b-it-q4\_K\_M         & Gemma           & 12 (M)                         & \texttt{q4\_K\_M}     & March 10, 2025        \\
        \texttt{gemma\_12b\_f} & gemma3:12b-it-fp16             & Gemma           & 12 (M)                         & \texttt{fp16}         & March 10, 2025        \\
        \texttt{gemma\_27b}    & gemma3:27b-it-q4\_K\_M         & Gemma           & 27 (L)                         & \texttt{q4\_K\_M}     & March 10, 2025        \\
        \texttt{llama\_3b}     & llama3.2:3b-instruct-q4\_K\_M  & Llama           & 3 (S)                          & \texttt{q4\_K\_M}     & September 25, 2024    \\
        \texttt{llama\_8b}     & llama3.1:8b-instruct-fp16      & Llama           & 8 (M)                          & \texttt{fp16}         & July 23, 2024         \\
        \texttt{llama\_8b\_f}  & llama3.1:8b-instruct-q4\_K\_M  & Llama           & 8 (M)                          & \texttt{q4\_K\_M}     & July 23, 2024         \\
        \texttt{llama\_70b}    & llama3.3:70b-instruct-q4\_K\_M & Llama           & 70 (L)                         & \texttt{q4\_K\_M}     & December 7, 2024      \\
        \texttt{qwen\_3b}      & qwen2.5:3b-instruct-q4\_K\_M   & Qwen            & 3 (S)                          & \texttt{q4\_K\_M}     & September 19, 2024    \\
        \texttt{qwen\_7b}      & qwen2.5:7b-instruct-q4\_K\_M   & Qwen            & 7 (M)                          & \texttt{q4\_K\_M}     & September 19, 2024    \\
        \texttt{qwen\_7b\_f}   & qwen2.5:7b-instruct-fp16       & Qwen            & 7 (M)                          & \texttt{fp16}         & September 19, 2024    \\
        \texttt{qwen\_72b}     & qwen2.5:72b-instruct-q4\_K\_M  & Qwen            & 72 (L)                         & \texttt{q4\_K\_M}     & September 19, 2024    \\
        \hline
    \end{tabular}
\end{table*}

\subsection{Prompt Design} \label{sec:methods_prompts}

The concrete prompt templates used in our experiments are detailed later in Section~\ref{sec:experiments} (Experimental Setup) and in Appendices~\ref{sec:appendix_prompt_impacts},~\ref{sec:appendix_prompt_relevance}, and~\ref{sec:appendix_prompt_locations}. In this section we describe the overall design principles and prompt engineering techniques.

To construct the base prompt template, we analyzed the drought-impact extraction task and identified the specific elements to be extracted. These elements were then expressed as explicit natural language instructions to reduce ambiguity and improve reproducibility. The resulting template includes:

\begin{itemize}
    \item \textbf{Task variables}, including the name of the climate event and its associated impact categories.
    \item \textbf{Input placeholders} for the news article’s content, such as the headline, body, and publication date, which may provide contextual information for extraction.
    \item \textbf{Role prompting strategy}, where the LLM is instructed to act as ``an expert in environmental analysis.'' This technique has been shown to improve performance by aligning model behavior with the target domain~\cite{chuNavigateEnigmaticLabyrinth2024,whitePromptPatternCatalog2023a}.
\end{itemize}

Although the input articles are written in Spanish, we formulate all prompts in English. This design choice is motivated by evidence that LLMs trained primarily on English corpora perform more reliably when prompted in English~\cite{fuPolyglotPromptMultilingual2022}. The final template was refined iteratively through empirical testing and meta prompting~\cite{reynoldsPromptProgrammingLarge2021,zhouLargeLanguageModels2023} on a small independent validation set to ensure consistency and robustness across different LLM architectures and sizes.

To improve performance beyond the base prompt, we evaluate four prompt engineering techniques commonly used in LLM-based task design:
\begin{itemize}
    \item \textbf{Summarization} (\texttt{SUM}): We first perform a separate LLM call to generate a summary of the news article. The resulting summary, rather than the full article, is then inserted into the extraction prompt. This strategy aims to reduce prompt length and eliminate irrelevant or superfluous content~\cite{liCompressingContextEnhance2023,muthukumarFrameworkAnalyzingSummarizing2024}.
    \item \textbf{Chain of Thought} (\texttt{CoT}): We append the instruction ``Reason step by step and explain your reasoning before giving the final answer'' to the base prompt. Zero-shot CoT techniques encourage intermediate reasoning steps, which have been shown to improve factual accuracy in complex tasks~\cite{kojimaLargeLanguageModels2023}.
    \item \textbf{Self-Criticism} (\texttt{SC}): Inspired by Self-Refine~\cite{madaanSELFREFINEIterativeRefinement2023}, we introduce a second LLM call that prompts to review its initial response. This self-critique may lead to more robust extractions by allowing the model to correct or refine prior outputs~\cite{huangLargeLanguageModels2023}.
    \item \textbf{Impact Descriptions} (\texttt{DESC}): Instead of relying on short label names for impact categories, we incorporate detailed natural language descriptions directly into the base prompt template. This approach aims to enhance the model’s ability to disambiguate overlapping or nuanced categories, at the cost of additional effort required from the researcher or system user to define these descriptions and an increase of the prompt's lenght.
\end{itemize}

Summarization and self-criticism require sequential LLM calls. They are implemented through dynamic prompt chaining, where the output of one step is parsed and embedded into the prompt for the next~\cite{yangMultistepIterativeAutomated2024}. Although this approach increases execution time, it enables the model to focus more effectively on intermediate subtasks, potentially leading to improved overall extraction performance.

\subsection{Structured Output Parsing} \label{sec:methods_parsing}

The LLM output must be converted to a structured format to enable downstream processing and evaluation. Our approach uses a schema that specifies the fields to be extracted (e.g., \texttt{event\_type}, \texttt{impacts}, \texttt{locations}) and serves as the blueprint for both guiding the LLM output and parsing it after generation.

We explore two distinct methods for instructing the model to generate its output in a structured JSON format, and denote this configuration variable as \texttt{JGEN}.

\begin{itemize}
    \item \textbf{Format-enforcing prompts} (\texttt{JGEN = prompt}): In this approach, the prompt explicitly provides JSON format instructions for the LLM to follow. These instructions are derived from the extraction schema and specify the expected structure, required fields, and valid data formats~\cite{baiSchemaDrivenInformationExtraction2023,luLearningGenerateStructured2025}.
    \item \textbf{Tool Calling Method} (\texttt{JGEN = tool}): For LLM APIs that support tool calling, we bind the extraction schema as a tool that the model can invoke~\cite{schickToolformerLanguageModels2023,heAchievingToolCalling2024}. The LLM is instructed, via system message or API configuration, to return a tool call using the schema as its signature. This enables the model to produce natively structured output that can be parsed reliably without relying on prompt adherence to syntax.
\end{itemize}

In addition to the method used to generate the structured output, we also vary the response parsing strategy. This determines when structuring takes place in the pipeline, as controlled by the configuration variable \texttt{RPARSE}:

\begin{itemize}
    \item \textbf{Single-Step Parsing} (\texttt{RPARSE = False}): The model is instructed to extract and structure the information in a single LLM call. This is the most efficient method but also the most error-prone, as it involves the extraction and the response formatting task.
    \item \textbf{Two-Step Parsing} (\texttt{RPARSE = True}): The extraction and formatting steps are decoupled. The model first performs information extraction in free-form text. A second LLM call is then used to transform this unstructured output into the desired JSON format. This method increases execution time but often improves formatting reliability, especially when models struggle to produce well-formed JSON in one step.
\end{itemize}

\subsection{CienaLLM Framework} \label{sec:methods_cienallm}

We have developed CienaLLM, an open-source Python framework for experimentation with LLM-based information extraction and is available at \url{https://github.com/lcsc/ciena_llm}. CienaLLM implements all the prompting strategies and response parsing methods discussed in Sections~\ref{sec:methods_llm}--~\ref{sec:methods_parsing}. The framework leverages LangChain\footnote{\url{https://www.langchain.com/}} to orchestrate prompt construction, LLM calls, output parsing, and schema enforcement. Parsing is implemented using Pydantic\footnote{\url{https://docs.pydantic.dev/latest/}}, which validates whether the generated output conforms to this structure. It integrates with Ollama to support local inference of open-weight models via llama.cpp\footnote{\url{https://github.com/ggml-org/llama.cpp}}.

CienaLLM is programmatically configurable in Python, allowing users to define (i) an extraction schema, (ii) a base prompt template dynamically populated with task information and additional prompt techniques, (iii) the LLM backend, and (iv) optional multi-step inference through additional LLM calls such as summarization or response parsing. All components are implemented as extensible modules, and configurations can be specified in a YAML file to enable reproducible and parameterized experiments. This modular design makes CienaLLM well suited for rapid experimentation in low-resource settings and across domains.

\begin{figure*}[!t]
    \centerline{\includegraphics[width=\textwidth]{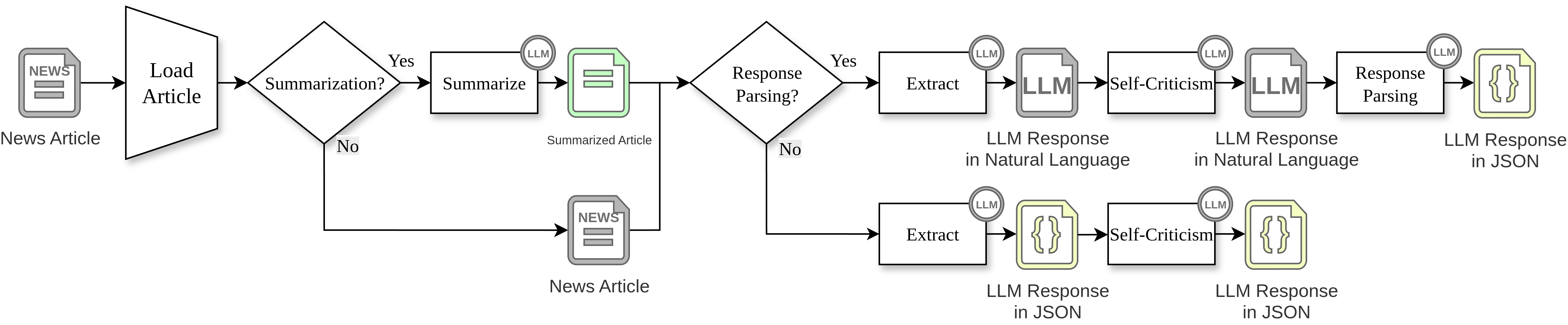}}
    \caption{Main components of the CienaLLM pipeline.}
    \label{fig:method_cienallm}
\end{figure*}

Figure~\ref{fig:method_cienallm} illustrates the main steps of the CienaLLM pipeline. A news article is ingested and parsed for metadata such as headline, body, and publication date. If enabled, a summarization step performs an initial LLM call. The resulting prompt is sent to the LLM to extract information, which may be returned in plain text or structured JSON. Optionally, a second LLM call reformats the response into structured output. A self-criticism step can further refine the answer. The final result is saved as a structured JSON object and exported in CSV format for evaluation. It includes robust logging and error handling mechanisms to ensure traceability and reliability: malformed outputs and JSON parsing failures are automatically detected and logged, and execution times for each processing step are recorded to facilitate performance diagnostics.

\section{Experimental Setup} \label{sec:experiments}

We design a series of experiments to evaluate the ability of open-weight LLMs to extract structured information of drought events from news articles. The main experiment explores how model family, size, quantization, and prompt engineering techniques affect drought impact extraction performance. In addition to this core task, we conduct two additional experiments: one to detect whether a news article mentions a drought event, and another to identify the geographic locations affected by drought events. Based on the validation results of the primary drought impact extraction task, we select three representative configurations capturing different points of the performance–efficiency trade-off. These profiles are then used for the final test evaluation of the core task and reused in the two secondary tasks.

\subsection{Drought Impact Extraction} \label{sec:experiments_impacts}

The primary goal of this evaluation is to assess the performance of large language models in extracting drought impacts from Spanish news articles. We use CienaLLM to run and evaluate the different model configurations, prompting strategies, and parsing techniques introduced in Section~\ref{sec:methods}. The task is formulated as a multi-label classification problem, where the system must determine which of the following impact categories are present in each article: agriculture, livestock, hydrological resources, and energy.  We evaluate model performance on the DID introduced in Section~\ref{sec:datasets_did} using both the validation and test splits.

To evaluate the impact of model choice and prompting strategies on extraction performance, we conduct an exhaustive exploration of all possible combinations of the 12 LLMs analyzed (Section~\ref{sec:methods_llm}), the 4 prompt design options (Section~\ref{sec:methods_prompts}), and whether parsing is performed (Section~\ref{sec:methods_parsing}), making a total of 384 configurations. This exhaustive experimental design allows us to systematically analyze the individual and combined effects of each factor on extraction performance. The \texttt{JGEN} variable is fixed to \texttt{prompt} in all experiments to ensure consistency, as the \texttt{gemma} models do not support tool-calling in Ollama.

We configure the CienaLLM framework with a task-specific prompt template and a schema that defines the structured output. The prompt is dynamically assembled from a base prompt template containing the core task instructions and placeholders for the article's headline and body. The activated experimental flags determine its final form: \texttt{SUM} replaces the full article with a summary generated in an additional LLM call, \texttt{DESC} inserts natural language definitions of each impact category in the base prompt, \texttt{CoT} appends an instruction to the base prompt to reason step by step before answering, and \texttt{SC} triggers a second LLM call to review and correct the initial output. Finally, When \texttt{RPARSE = false}, the model is instructed to output JSON directly within the main extraction prompt; and when \texttt{RPARSE = true}, a second LLM call reformats the initial free-text extraction into JSON using a dedicated response parsing prompt. LLM responses are expected to follow a structured JSON format consisting of four binary fields as defined in the extraction schema (\texttt{agriculture}, \texttt{livestock}, \texttt{hydrological\_resources}, and \texttt{energy}). A complete set of prompt templates, including examples for each component, as well as the complete schema definition, are provided in Appendix~\ref{sec:appendix_prompt_impacts}.

To ensure reproducibility and consistency across runs, all LLM generations in the CienaLLM pipeline are performed with temperature set to 0, enforcing a greedy decoding strategy in which the model deterministically selects the most probable next token at each step. In addition, a fixed seed is set in the Ollama backend to ensure the same output is obtained across different machines and environments. To prevent excessively long generations or infinite loops, a maximum output length of 2,048 tokens is enforced for each generation call. The context window of the models is set at a limit of 32,768 tokens, which accommodates all prompt variants and news articles in the datasets.

We evaluate model performance on the drought impact extraction task using standard multi-label classification metrics (accuracy, precision, recall, and F1 score), computed as micro-averages across the four impact categories~\cite{powersEvaluationPrecisionRecall2020a,sokolovaSystematicAnalysisPerformance2009,zhangReviewMultiLabelLearning2014,Tsoumakas2010}. These metrics are calculated only on successfully parsed outputs, ensuring that extraction accuracy is assessed independently of formatting issues. To quantify reliability, we report the parsing error rate, defined as the percentage of outputs that could not be converted into the expected JSON format according to the extraction schema. Efficiency is measured as execution time per article, reflecting the wall-clock time to process one article through the full extraction pipeline, excluding model loading and warm-up.

For most validation-set results in the impact extraction task, we report three core metrics: F1 score, parsing error rate, and execution time per article. Together, these capture accuracy, reliability, and efficiency for comparing configurations during exploratory analysis. For the final test-set evaluation, we additionally report accuracy, precision, and recall alongside these core metrics.

To determine whether specific design choices significantly affect performance and efficiency, we conduct a series of non-parametric statistical tests across the full factorial space of experiments. We apply the Wilcoxon Signed-Rank test~\cite{wilcoxonIndividualComparisonsRanking1945}, pairing configurations that differ only in the factor under analysis while keeping all other variables constant. Tests are run on every evaluation metric, using only complete pairs. Statistical significance is assessed at $\alpha = 0.05$ with Bonferroni correction~\cite{neymanUseInterpretationCertain1928,armstrongWhenUseOnferroni2014} for multiple comparisons.

In addition to pairwise significance testing, we conduct a multi-objective Pareto-front analysis~\cite{fukazawaParetoFrontAnalysis2023,liuPowerParetoFront2025} to characterize trade-offs between extraction performance and computational cost, treating each configuration as a point in the space defined by F1 score and execution time per article. From the results of the validation set, we identify three representative configurations from the Pareto front: \textit{Best-F1}, \textit{Efficient}, and \textit{Fastest} (see Section~\ref{sec:results_tradeoff}). These capture different points along the trade-off curve and are reused for the evaluation of the supplementary tasks.

\subsection{Drought Relevance Classification} \label{sec:experiments_relevance}

In addition, we asses the ability of large LLMs to detect whether a news article provides drought-relevant information, formulated as a binary classification. The evaluation is performed on the DRD test split, using the three representative configurations from the primary drought impact extraction task. A simplified prompt, derived from the base extraction template, asks the model to determine whether the article reports on a drought event, returning a JSON object with a single boolean field (\texttt{"drought"}). See the Appendix~\ref{sec:appendix_prompt_relevance} for further information on the prompts and schema.

\subsection{Drought Impact Location Extraction} \label{sec:experiments_locations}

Lastly, we evaluate the ability of our methodology to extract the geographic scope of drought impacts, specifically identifying the Spanish provinces mentioned or implied in news articles. The evaluation uses the DILD dataset and the three selected representative configurations.

The prompt asks the model to identify all Spanish provinces affected by drought according to the article. The expected output is a JSON object with the field \texttt{"provinces"} containing a list of province names. The prompt and corresponding schema are included in the Appendix~\ref{sec:appendix_prompt_locations}. After model inference, predicted province names are post-processed and normalized to match canonical Spanish province names. This step accounts for possible variations introduced by the model, including alternative spellings, punctuation differences, and the use of regional languages.

\subsection{Infrastructure and Execution Environment} \label{sec:experiments_infra}

Although, initially tested on a consumer-grade NVIDIA RTX A4000 GPU (16 GB), the extensive factorial experiment with 384 configurations required a scalable infrastructure. We utilized the Galicia Supercomputing Center (CESGA)\footnote{\url{https://www.cesga.es}}, specifically their GPU cluster with 64 nodes, each featuring two NVIDIA A100 GPUs (40 GB each). The CienaLLM framework (version v0.3.0\footnote{\url{https://github.com/lcsc/ciena_llm/releases/tag/v0.3.0}}) was used to execute the extraction pipeline on an Ollama server on-premises. All parameters and outputs were systematically logged for traceability and reproducibility, enabling efficient, controlled exploration of model and prompt combinations.

\section{Results} \label{sec:results}

\subsection{Response Parsing and Reliability} \label{sec:results_rparse}

Experiments without response parsing exhibited failures across models (e.g., trailing commas, missing required fields), causing loss of valid outputs. Averaged across all configurations and models, the parsing error rate without response parsing (\texttt{RPARSE = False}) reached 4.0\%, compared to only 0.7\% with response parsing (\texttt{RPARSE = True}) (see Table~\ref{tab:appendix_results_rparse}).

\begin{figure*}[!t]
    \centering
    \includegraphics{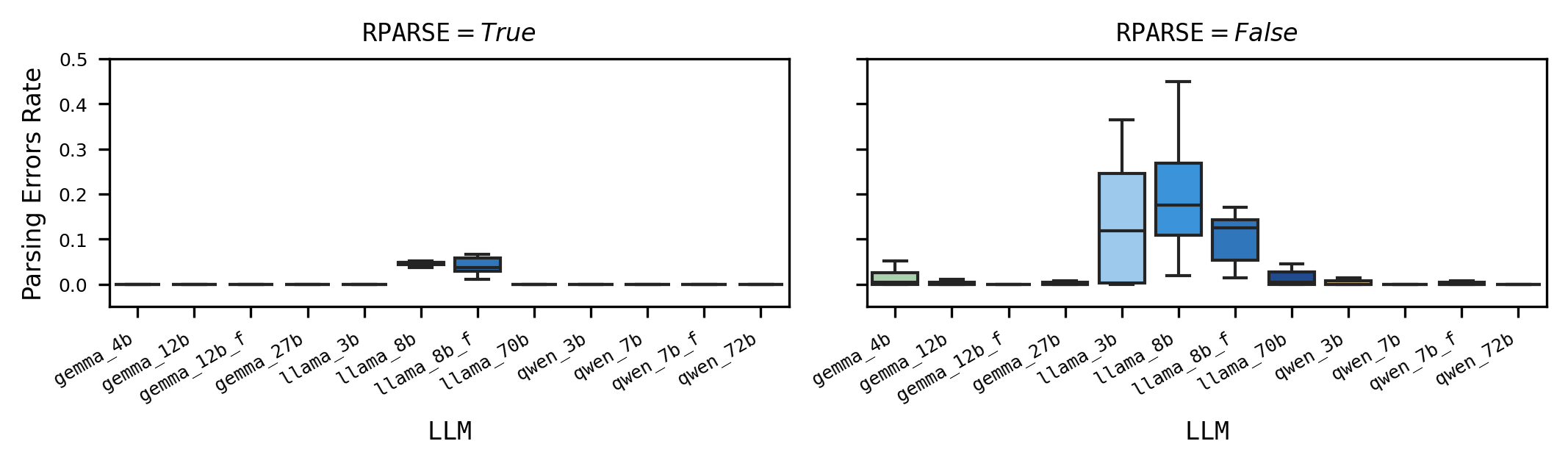}
    \caption{Parsing error rates by model (\texttt{LLM}) with response parsing (\texttt{RPARSE}) enabled (left) and disabled (right).Without response parsing, several models, particularly the smaller Llama variants, exhibit high and variable error rates. In contrast, response parsing eliminates such failures across nearly all models.}
    \label{fig:results_rparse}
\end{figure*}

At the model level (see Table~\ref{tab:model_metrics_response_parsing}), the impact of \texttt{RPARSE} was uneven. Some models, such as \texttt{llama\_8b}, failed to produce valid JSON for nearly 19\% in some configurations, whereas \texttt{qwen} and \texttt{gemma} models already showed near-perfect formatting ($<1\%$ errors). \texttt{RPARSE} reduced average parsing error rates for the most problematic ones i.e. \texttt{llama\_8b} and \texttt{llama\_3b}, from 13\% and 19\% respectively to below 4.5\% in the least reliable configurations (see Figure~\ref{fig:results_rparse}).

However, parsing did not affect extraction accuracy. The F1 scores, computed only on successfully parsed outputs, remained almost identical for both settings, averaging $0.803 \pm 0.061$ without response parsing and $0.808 \pm 0.048$ with it (adjusted $p=1.0$). The main drawback of enabling RPARSE is a moderate increase in execution time, from $10.2 s \pm 10.6$ to $16.1 s \pm 15.0$ per article, but this cost is outweighed by its reliability gains. The substantial loss of information caused by parsing errors in some configurations makes the results not directly comparable to those of more reliable models. Consequently, to ensure that all models are evaluated under equally reliable conditions, we exclude from the analysis configurations that lack response parsing.

\subsection{Model-Level Performance Analysis} \label{sec:results_model}

\begin{figure}[htbp]
    \centering
    \includegraphics{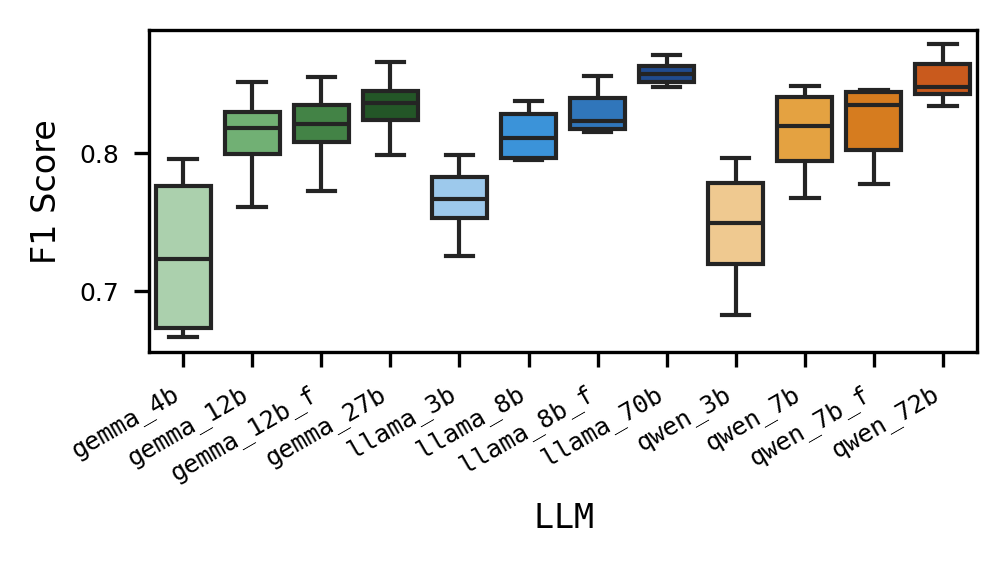}
    \caption{Distribution of F1 scores across prompt configurations for each mode (\texttt{LLM}). Larger models generally achieve higher and more stable performance.}
    \label{fig:results_model}
\end{figure}

Figure~\ref{fig:results_model} shows performance metrics for each model across configurations. Larger models exhibit both high average F1 and low variance, reflecting strong and stable performance across prompt configurations. In contrast, smaller models show lower scores and greater variance, with performance ranging widely depending on the configuration. This variability reflects each model's prompt sensitivity, that is, its susceptibility to changes in prompt formulation.

On average, \texttt{llama\_70b} achieves the highest F1 score (0.858), followed by \texttt{qwen\_72b} (0.852) and \texttt{gemma\_27b} (0.833), all with perfect parsing reliability (see Table~\ref{tab:appendix_results_model}). These results confirm the advantage of scale in achieving both accuracy and consistency. Smaller models like \texttt{gemma\_4b} and \texttt{qwen\_3b} trail behind in average performance and are more affected by prompt changes.

The best configuration for each model (see Table~\ref{tab:appendix_results_model_top}) shows that most top-performing setups include \texttt{DESC}, often combined with \texttt{CoT}. For instance, \texttt{qwen\_72b} reaches the highest F1 overall (0.878) with \texttt{CoT + DESC}, and \texttt{gemma\_27b} reaches 0.865 using \texttt{SUM + CoT + DESC}. For \texttt{gemma}, \texttt{SUM} appears particularly effective and are featured in the best configuration of every model of the family. However, smaller models like \texttt{llama\_3b} or \texttt{qwen\_3b} achieve their best scores without prompt enhancements.

Full-precision (\texttt{fp16}) models consistently outperform their quantized (\texttt{q4\_K\_M}) counterparts by small but statistically significant margins: +0.008 F1 for \texttt{gemma}, +0.017 for \texttt{llama}, and +0.008 for \texttt{qwen} (see Table~\ref{tab:appendix_results_quantization}). However, execution times are reduced by a 30\%--40\% while parsing reliability remains unaffected.

\subsection{Effect of Prompt Strategies Across Models} \label{sec:results_prompt}

\begin{figure}[htbp]
    \centering
    \includegraphics{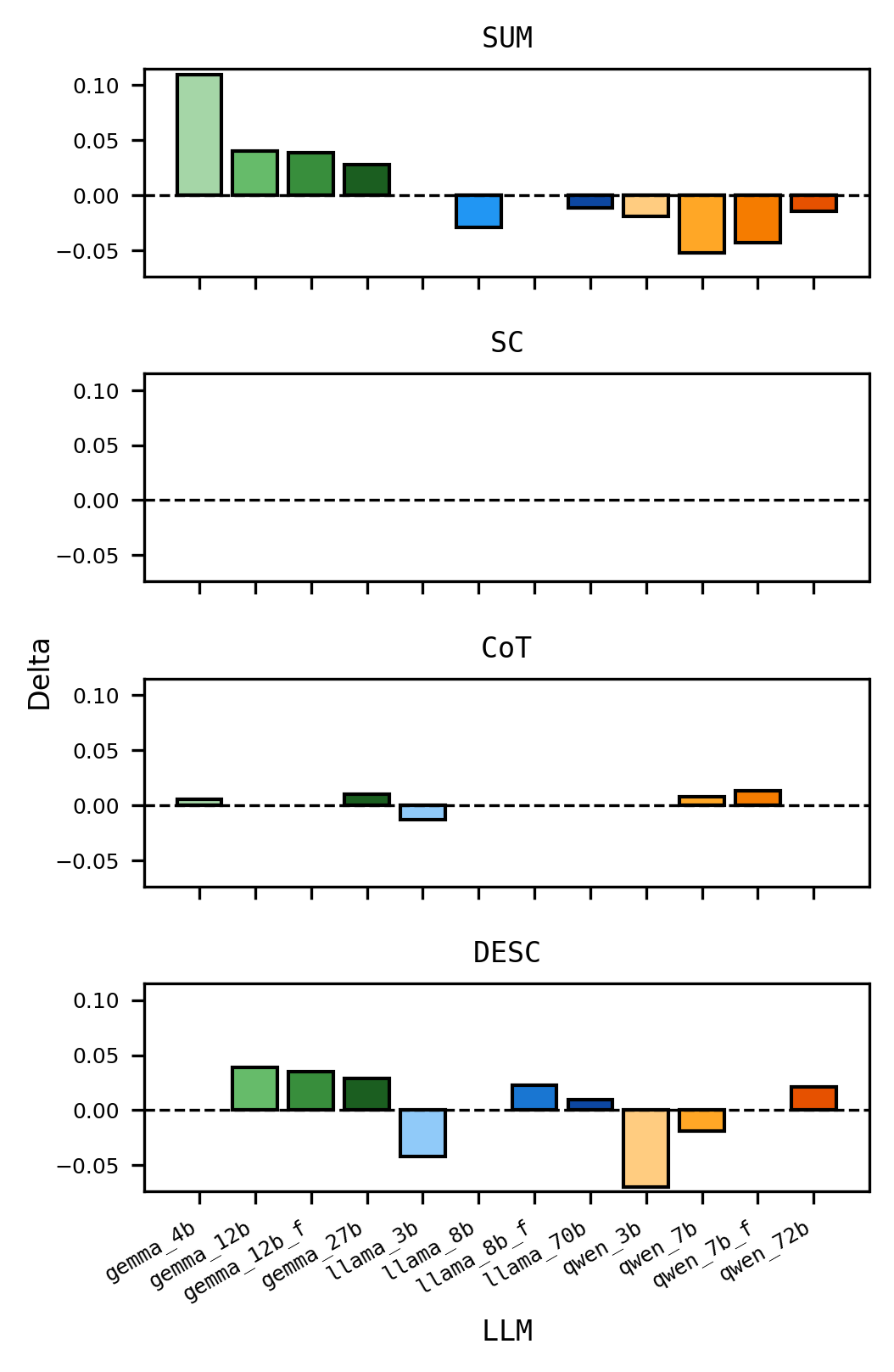}
    \caption{Effect of prompt strategies on F1 scores by model (\texttt{LLM}). Bars show the change in mean F1 ($\Delta$) relative to the base prompt for Summarization (\texttt{SUM}), Self-Criticism (\texttt{SC}), Chain of Thought (\texttt{CoT}), and Impact Descriptions (\texttt{DESC}), showing only statistically significant changes.}
    \label{fig:results_prompt}
\end{figure}

We evaluate the impact of each prompt strategy on model performance and show the results in Table~\ref{tab:appendix_results_prompt}. On average across all models, performance differences are small, indicating no universally dominant strategy. Execution time increases notably for strategies requiring extra LLM calls: SUM (+4.1 s/article), SC (+7.8 s/article), and CoT (+2.5 s/article), while parsing error rates remain largely unaffected. Clearer patterns emerge when analyzing at the model (see Figure~\ref{fig:results_prompt} and Table~\ref{tab:appendix_results_prompt_model}) and family level (see Table \ref{tab:appendix_results_prompt_family}):

\begin{itemize}
    \item \textbf{Summarization} (\texttt{SUM}): Consistently improves performance in all \texttt{gemma} models, with statistically significant gains averaging +0.054, but has a negative effect on \texttt{llama} and \texttt{qwen} models, particularly in the medium-size range. This suggests \texttt{gemma} benefits from shorter, focused inputs, whereas the others rely more on full-article context.
    \item \textbf{Self-criticism} (\texttt{SC}): Provides no measurable gains. For many \texttt{llama} and \texttt{qwen} models, responses were identical with and without \texttt{SC}. Logs confirm that the instruction was ignored, suggesting the models tended to rely on their initial outputs, adding latency without benefit.
    \item \textbf{Chain of Thought} (\texttt{CoT}): Yields only modest improvements ($\pm0.01$), significant in a few medium/large models such as \texttt{gemma\_27b} and \texttt{qwen\_7b\_f}. Larger checkpoints appear able to exploit explicit reasoning, whereas smaller ones lack sufficient capacity.
    \item \textbf{Impact Descriptions} (\texttt{DESC}): Slightly improves results for larger models (e.g., +0.009 for \texttt{llama\_70b} and +0.021 for \texttt{qwen\_72b}), while smaller models like \texttt{qwen\_3b} and \texttt{llama\_3b} achieve better results without the extra descriptive text. This indicates that richer label guidance pays off only when model capacity is sufficient.
\end{itemize}

These findings are consistent with previous section results (see Table~\ref{tab:appendix_results_model_top}): the best-performing configuration for most models include \texttt{DESC}, \texttt{SUM} appears only for \texttt{gemma}, \texttt{CoT} helps about half the time, and small models often use no prompt enhancements at all.

\subsection{Efficiency vs Performance Trade-Offs} \label{sec:results_tradeoff}

\begin{figure}[htbp]
    \centering
    \includegraphics{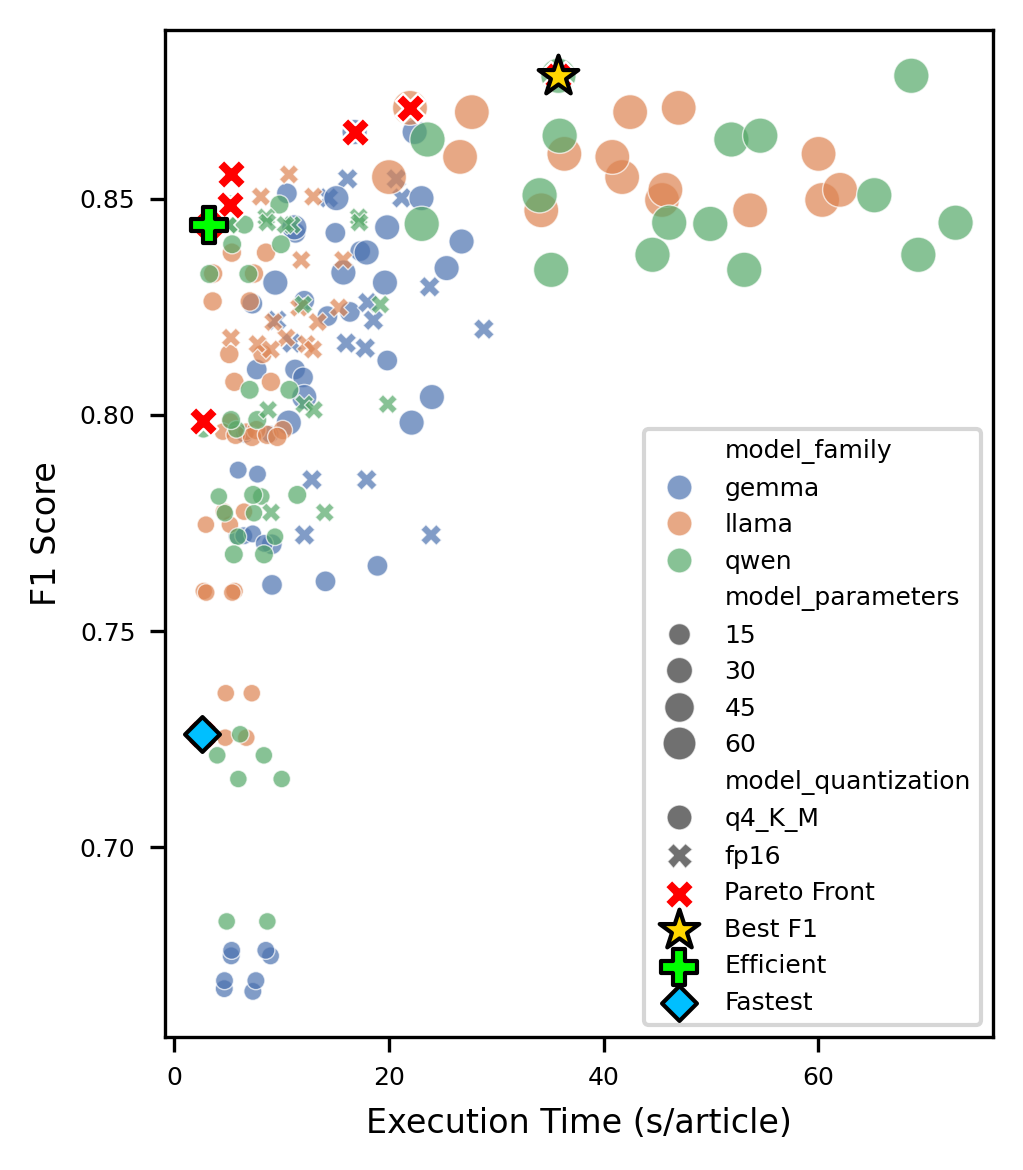}
    \caption{Multi-objective Pareto-front analysis of all evaluated configurations, showing the trade-off between F1 score and execution time per article. Pareto-optimal configurations are marked with red crosses, and three representative setups, \textit{Best-F1}, \textit{Efficient}, and \textit{Fastest}, are highlighted.}
    \label{fig:results_tradeoff}
\end{figure}

To better understand the trade-off between extraction performance and computational cost, we conducted a multi-objective Pareto-front analysis using the F1 score and the execution time per article. Figure~\ref{fig:results_tradeoff} visualizes all 384 evaluated configurations, highlighting those on the Pareto front, that is, configurations that cannot be outperformed simultaneously in both accuracy and efficiency. The results shows a clear performance–efficiency gradient: large models dominate the high-F1 region but incur high inference costs, while small and medium models achieve competitive efficiency, with some approaching the accuracy of much larger models.

From the Pareto front, we highlight three representative configurations that capture the main trade-offs: \textit{Best-F1}, \textit{Efficient}, and \textit{Fastest} (see Table~\ref{tab:appendix_results_tradeoff}). The \textit{Best-F1} configuration (\texttt{qwen\_72b + CoT + DESC + PARSE}) corresponds to the setting with the highest overall F1 score (0.878), though at a cost of 35.8 s/article. The \textit{Fastest} setup (\texttt{qwen\_3b + PARSE}) represents the minimum execution time, processing an article in only 2.6 s with moderate accuracy (F1 = 0.726). The \textit{Efficient} profile (\texttt{qwen\_7b + DESC + PARSE}) was explicitly chosen to represent a balanced compromise, achieving F1 = 0.844 at just 3.2 s/article.

These configurations align with our earlier findings on prompt strategies. The \textit{Best-F1} model benefited from richer guidance through \texttt{CoT} and \texttt{DESC}, while omitting summarization, which was ineffective outside \texttt{gemma}. The \textit{Fastest} configuration used no enhancements, consistent with the tendency of small models to ignore or degrade under additional prompting. The \textit{Efficient} setup leveraged \texttt{DESC} effectively, confirming its value as the most cost-efficient strategy when model capacity is sufficient. In terms of resources, the computational requirements scale sharply with model size: the \texttt{qwen\_72b} model demands high GPU resources, the \texttt{qwen\_3b} model could run on the CPU. The \texttt{qwen\_7b} setup offers a practical compromise, delivering accuracy near larger LLMs on consumer-grade GPUs.

Altogether, the three selected configurations capture the performance–efficiency spectrum and will be used as reference points in the following sections to evaluate high-accuracy, cost-effective, and resource-constrained setups.

\subsection{Drought Impact Extraction Performance} \label{sec:results_impacts}

We evaluate the three selected configurations on drought impact extraction using the DID test split. Overall, \textit{Best-F1} achieves the strongest results (F1 = 0.873, recall = 0.911, and precision = 0.837), but at the cost of long runtimes (over 36 s/article). \textit{Fastest} processes articles in just 2.47 s, but its performance drops (F1 = 0.646), particularly due to poor recall (0.538). \textit{Efficient} reaches a competitive performance (F1 = 0.808) with much faster inference (3.47 s/article), offering a solid balance between accuracy and efficiency. None of the three configurations exhibited parsing errors. These results are summarized in Table~\ref{tab:results_impacts}.

Breaking down performance by impact category (see Table~\ref{tab:results_impacts_impacts}) shows that \textit{Best-F1} leads on Agriculture (0.868), Livestock (0.966), and Hydrological Resources (0.824), while \textit{Efficient} achieves the best score for Energy (0.970). \textit{Fastest} underperforms across categories confirming the limitations of smaller models in capturing complex or implicit impacts.

To ensure a fair comparison with SeqIA~\cite{lopez-otalSeqIAPythonFramework2025}, we use the E2E dataset for evaluation, which was specifically created for this purpose~\cite{lopezotalSeqIAAnnotatedDroughtrelated2025}. DID is excluded from this comparison to avoid training overlap, as specified on Section~\ref{sec:datasets_did}. CienaLLM’s \textit{Best-F1} configuration surpasses SeqIA (F1 = 0.782 vs. 0.769) but at $\sim20\times$higher per-article latency (30.07 s vs. 1.54 s/article). The \textit{Efficient} and \textit{Fastest} profiles approach SeqIA’s runtime (2.54 s and 1.81 s/article) but with lower performance (F1 = 0.645 and 0.452, respectively). These results are presented in Table~\ref{tab:results_impacts_e2e}.

\begin{table*}[!t]
    \centering
    \caption{Drought impact extraction performance on the test split of the DID dataset for the three representative configurations.}
    \label{tab:results_impacts}
    \small
    \begin{tabular}{|l|S[table-format=1.3]|S[table-format=1.3]|S[table-format=1.3]|S[table-format=1.3]|S[table-format=1.3]|S[table-format=1.3]|}
        \hline
        \textbf{Configuration} & \textbf{Accuracy} & \textbf{Precision} & \textbf{Recall} & \textbf{F1 Score} & \textbf{Parsing Error Rate} & \textbf{Exec. Time (s/article)} \\
        \hline
        \textit{Best-F1}       & 0.684             & 0.837              & 0.911           & 0.873             & 0.000                       & 36.432                          \\
        \textit{Fastest}       & 0.436             & 0.810              & 0.538           & 0.646             & 0.000                       & 2.468                           \\
        \textit{Efficient}     & 0.598             & 0.832              & 0.785           & 0.808             & 0.000                       & 3.320                           \\
        \hline
    \end{tabular}
\end{table*}

\begin{table*}[!t]
    \centering
    \caption{Per-impact drought impact extraction performance on the test split of the DID dataset for the three representative configurations.}
    \label{tab:results_impacts_impacts}
    \small
    \begin{tabular}{|l|S[table-format=1.3]|S[table-format=1.3]|S[table-format=1.3]|S[table-format=1.3]|}
        \hline
        \textbf{Configuration} & \textbf{Agriculture} & \textbf{Livestock} & \textbf{Hydrological Resources} & \textbf{Energy} \\
        \hline
        \textit{Best-F1}       & 0.868                & 0.966              & 0.824                           & 0.914           \\
        \textit{Fastest}       & 0.634                & 0.542              & 0.674                           & 0.737           \\
        \textit{Efficient}     & 0.780                & 0.893              & 0.746                           & 0.970           \\
        \hline
    \end{tabular}
\end{table*}

\begin{table*}[htbp]
    \centering
    \caption{Drought impact extraction performance on the E2E dataset for CienaLLM's three representative configuration and SeqIA.}
    \label{tab:results_impacts_e2e}
    \small
    \begin{tabular}{|l|S[table-format=1.3]|S[table-format=1.3]|S[table-format=1.3]|S[table-format=1.3]|S[table-format=1.3]|S[table-format=1.3]|}
        \hline
        \textbf{Configuration} & \textbf{Accuracy} & \textbf{Precision} & \textbf{Recall} & \textbf{F1 Score} & \textbf{Parsing Error Rate} & \textbf{Exec. Time (s/article)} \\
        \hline
        \textit{Best-F1}       & 0.742             & 0.737              & 0.832           & 0.782             & 0.000                       & 30.070                          \\
        \textit{Fastest}       & 0.634             & 0.769              & 0.320           & 0.452             & 0.000                       & 1.811                           \\
        \textit{Efficient}     & 0.686             & 0.791              & 0.544           & 0.645             & 0.000                       & 2.540                           \\
        SeqIA                  & 0.788             & 0.735              & 0.806           & 0.769             & 0.000                       & 1.540                           \\
        \hline
    \end{tabular}
\end{table*}

\subsection{Drought Relevance Classification Performance} \label{sec:results_relevance}

Relevance classification results of CienaLLM's configuration and SeqIA on the DRD test split are summarized in Table~\ref{tab:results_relevance}. The \textit{Best-F1} configuration achieved the strongest overall performance (F1 = 0.968), while maintaining near-perfect parsing reliability (0.6\%). However, it was also the slowest, requiring 11.28 s per article. The \textit{Efficient} setup offered a strong trade-off, with F1 = 0.929, balanced precision (0.968) and recall (0.894), and perfect parsing reliability, at a much faster 1.74 s per article. The \textit{Fastest} profile processed articles in 1.36 s, but its accuracy degraded markedly (F1 = 0.779), especially due to lower recall (0.698), and its parsing error rate rose to 25\%. Compared to SeqIA (F1 = 0.961, execution time = 0.18 s/article), \textit{Best-F1} achieved nearly identical performance, but with higher latency ($\sim60\times$ faster). \textit{Efficient} remained competitive at $\sim10\times$ SeqIA's latency, while \textit{Fastest} underperformed both in speed ($\sim8\times$ slower) and accuracy. These results show that while CienaLLM can match or slightly surpass SeqIA in accuracy, efficiency remains a critical limitation.

\begin{table*}[!t]
    \centering
    \caption{Drought relevance classification performance on the test split of the DRD dataset for CienaLLM’s three representative configurations and SeqIA.}
    \label{tab:results_relevance}
    \small
    \begin{tabular}{|l|S[table-format=1.3]|S[table-format=1.3]|S[table-format=1.3]|S[table-format=1.3]|S[table-format=1.3]|S[table-format=2.2]|}
        \hline
        \textbf{Configuration} & \textbf{Accuracy} & \textbf{Precision} & \textbf{Recall} & \textbf{F1 Score} & \textbf{Parsing Error Rate} & \textbf{Exec. Time (s/article)} \\
        \hline
        \textit{Best-F1}       & 0.965             & 0.967              & 0.970           & 0.968             & 0.006                       & 11.282                          \\
        \textit{Efficient}     & 0.925             & 0.968              & 0.894           & 0.929             & 0.000                       & 1.738                           \\
        \textit{Fastest}       & 0.792             & 0.880              & 0.698           & 0.779             & 0.249                       & 1.355                           \\
        \textit{SeqIA}         & 0.957             & 0.957              & 0.965           & 0.961             & 0.000                       & 0.180                           \\
        \hline
    \end{tabular}
\end{table*}

\subsection{Drought Impact Location Extraction Performance} \label{sec:results_locations}

Table~\ref{tab:results_locations} summarizes the performance of the three selected configurations on drought impact location extraction. SeqIA results are not included here, as the system extracts all mentioned toponyms, whereas CienaLLM was prompted to extract only the provinces affected by drought, making the tasks not directly comparable.

\textit{Best-F1} achieved the strongest results, though performance remained modest (F1 = 0.465), a notable parsing error rate (2\%), and high inference cost (22.39 s/article). The \textit{Efficient} setup reached lower F1 (0.233) and suffered from unreliable parsing error rate (50\%). The \textit{Fastest} profile, while processing articles in just 1.37 s/article, produced an almost negligible F1 of 0.007. Across all configurations, precision remained moderate ($\sim0.75$), but recall was consistently poor (0.34--0.00), along with parsing issues. Overall, these results underscore both the difficulty of location extraction and the substantial performance gap across model scales.

\begin{table*}[!t]
    \centering
    \caption{Drought impact location extraction performance on the DIDL dataset for CienaLLM’s three representative configurations.}
    \label{tab:results_locations}
    \small
    \begin{tabular}{|l|S[table-format=1.3]|S[table-format=1.3]|S[table-format=1.3]|S[table-format=1.3]|S[table-format=1.3]|S[table-format=2.2]|}
        \hline
        \textbf{Configuration} & \textbf{Accuracy} & \textbf{Precision} & \textbf{Recall} & \textbf{F1 Score} & \textbf{Parsing Error Rate} & \textbf{Exec. Time (s/article)} \\
        \hline
        \textit{Best-F1}       & 0.540             & 0.734              & 0.340           & 0.465             & 0.020                       & 22.389                          \\
        \textit{Efficient}     & 0.400             & 0.762              & 0.138           & 0.233             & 0.500                       & 1.845                           \\
        \textit{Fastest}       & 0.330             & 0.750              & 0.004           & 0.007             & 0.620                       & 1.375                           \\
        \hline
    \end{tabular}
\end{table*}

\section{Discussion} \label{sec:discussion}

Structured extraction of climate impacts from news articles supports effective monitoring and understanding of the socio-economic consequences of extreme events. While supervised systems can achieve strong in-domain accuracy~\cite{sodogeAutomatizedSpatiotemporalDetection2023,lopez-otalSeqIAPythonFramework2025}, they require costly annotation and retraining whenever new tasks, hazards, or languages are introduced. Previous explorations of generative approaches have remained narrow, relying on sources like Wikipedia that are far less ambiguous and heterogeneous than news articles~\cite{liUsingLLMsBuild2024}. To our knowledge, no prior work has systematically evaluated open-weight LLMs for climate-impact extraction from news, nor compared across model families, sizes, precision regimes, and prompting strategies. Our study addresses this gap through the development of CienaLLM, a schema-guided GenIE framework that uses zero-shot prompting to deliver competitive accuracy without retraining, and flexibly adapts to evolving schemas. These features position CienaLLM as a practical and adaptable alternative for transforming news into structured datasets that enable systematic climate-risk monitoring and adaptation planning.

Building on this framework, our experiments highlight several factors that critically affect extraction performance and reliability. Response parsing proved essential, since without it some models often produced trivial JSON defects that compromised cross-model comparability. These effects are consistent with findings~\cite{luLearningGenerateStructured2025} that show that even recent LLMs struggle to generate valid JSON outputs across diverse schemas. An additional parsing step almost entirely eliminated such errors without reducing accuracy, so findings reported in the parsing-enabled setting remain valid in a no-parsing regime. Although this added latency, ensuring near-perfect parsability was essential; otherwise, valuable information from news articles would be lost.

We find a clear temporal trend in model reliability, as newer LLM releases generate valid JSON far more consistently. This improvement appears linked to recent training practices that emphasize structured output for tool use~\cite{wangSLOTStructuringOutput2025,quToolLearningLarge2025} together with the growing number of benchmarks that encourage schema compliance~\cite{caoStructEvalDeepenBroaden2024,gengJSONSchemaBenchRigorousBenchmark2025}. In contrast, older models remain far less reliable (see release dates in Table~\ref{tab:methods_llm}). We found no evidence that parsing success varied by article characteristics; instead, it depends primarily on each model’s ability to follow output instructions. As structured response methods continue to advance, the reliance on explicit response parsing may diminish, but remains indispensable for fair and reliable evaluation in our experiments.

As expected, one of the strongest patterns in our results is that larger models consistently outperform smaller ones, not only in peak accuracy but also in stability and robustness. Larger LLMs show much lower variance across prompt configurations, handling implicit or subtle mentions of impact more reliably. Smaller models, by contrast, are more sensitive to prompt design and show higher variance, having some prompt configurations approach the performance of their larger counterparts. Parameter count closely tracks F1 score and per-article runtime, while model family plays a comparatively minor role.

Precision strongly shapes efficiency, since in our experiments quantization reduced latency and memory demands substantially while incurring only modest accuracy penalties. Similar findings have been reported in broader evaluations of quantization techniques~\cite{jinComprehensiveEvaluationQuantization2024}. This balance makes quantized models especially attractive for deployment in settings with limited hardware resources. Additionally, studies on quantization~\cite{badshahQuantifyingCapabilitiesLLMs2024} show that larger models are more resilient to reductions in precision: even when quantized, many high-parameter models show only modest decreases in accuracy relative to full-precision versions. While our setup used one specific quantization regime, studies~\cite{kurticGiveMeBF162025,leeQrazorReliableEffortless2025} present alternative schemes that offer different trade-offs between speed, memory usage, and precision.

Since this study’s experiments, improved model releases have become available. For example, Qwen 3\footnote{\url{https://qwenlm.github.io/blog/qwen3/}} and Llama 4\footnote{\url{https://www.llama.com/models/llama-4/}}, released in April 2025. CienaLLM’s modular design allows updated models and quantization techniques to be integrated via simple configuration changes. Following empirical scaling laws~\cite{kaplanScalingLawsNeural2020}, which show performance scaling with model size, dataset size, and compute, we expect that overall performance of our methodology, in terms of accuracy, robustness, and efficiency, should continue to improve as newer models evolve.

Prompt interventions showed no universal recipe for performance gains, but they remain a valuable lever for adaptation. On average, their effects were modest yet highly dependent on model family and scale. Some models benefited from condensed inputs or richer label descriptions, consistent with previous findings~\cite{liuEffectsPromptLength2025}, whereas others relied on full context or ignored additional guidance. These results suggest that prompt design is best seen as a model- and task-specific hyperparameter. It is not a guaranteed accuracy booster, but rather a practical way to extract more reliability from limited models~\cite{sahooSystematicSurveyPrompt2025}. Importantly, this means that when resources are constrained, smaller models can approximate the performance of larger ones through carefully tuned prompt engineering, offering a flexible path to balance cost and accuracy.

Across impact extraction, relevance classification, and location identification, our results demonstrate that schema-guided GenIE with open-weight LLMs is a viable alternative to supervised baselines like SeqIA~\cite{lopez-otalSeqIAPythonFramework2025}. While SeqIA continues to dominate in throughput and efficiency, CienaLLM achieves competitive or superior accuracy for impact extraction and relevance classification, while uniquely supporting schema-flexible location extraction. Additionally, performance scales predictably with LLM generation, meaning that CienaLLM can seamlessly integrate future model releases to improve accuracy, stability, and efficiency without retraining. This adaptability positions CienaLLM and the GenIE approach as a practical alternative for climate-impact extraction.

On the drought impact extraction task, CienaLLM’s selected configurations achieve strong results on the DID dataset, but performance drops by 10--20 points on the E2E dataset~\cite{lopezotalSeqIAAnnotatedDroughtrelated2025}, even though the performance for \textit{Best-F1} and SeqIA is still matched. This gap is explained by differences between the datasets: DID contains only drought-related news and was designed specifically for impact extraction, whereas E2E evaluates the full pipeline, including drought relevance detection and therefore includes drought-unrelated articles. Internal validations restricted to drought-related E2E articles confirm that CienaLLM performs better in that setting, although still below DID levels. Beyond this, differences in dataset composition might contribute further to the gap: E2E exhibits broader thematic coverage, fewer impacts per article, and more imbalanced impact labels. Overall, DID is the more suitable benchmark for assessing impact extraction, while E2E remains the only option for a fair comparison with SeqIA.

For detecting drought-relevant articles, CienaLLM reaches accuracy comparable to the supervised SeqIA classifier, but at a much higher inference cost, making it unsustainable for large-scale preprocessing. In such scenarios, faster approaches are preferable. Supervised classifiers, when annotated data is available, provide reliable filtering at low latency, while keyword matching offers a simple and inexpensive alternative when labeled datasets are lacking, albeit at lower accuracy~\cite{lopez-otalSeqIAPythonFramework2025}. For new events without an existing classifier, keyword filtering may be the most practical option. A hybrid strategy, which uses lightweight supervised or keyword-based filtering to pre-select relevant articles and then applies CienaLLM for detailed impact extraction, offers a balanced trade-off between efficiency, flexibility, and accuracy.

The difficulty of impacted location extraction reflects a broader challenge in climate information extraction, namely the need to move from unstructured mentions in news articles to precise geographic entities. Previous work~\cite{lopez-otalSeqIAPythonFramework2025} has often relied on named entity recognition (NER) to identify place names, followed by post-processing or gazetteer matching to disambiguate and geolocate them. Yet, ambiguity persists: When an article mentions a river, does the impact extend to all provinces it crosses? When it refers to ``the northeast of the peninsula,'' how should that region be delineated? More fundamentally, translating text into mappable units requires choosing an appropriate spatial scale. Drought typically affects broad areas, which we operationalize at the provincial level, but other hazards may demand finer (e.g., municipal) resolutions. Defining these boundaries is inherently complex, as climate impacts rarely align with administrative divisions.

Our results highlight this challenge: high precision but low recall indicates that models are cautious about overgeneralization, yet often fail to capture implicitly referenced locations. By extracting impacted rather than merely mentioned locations, CienaLLM produces spatially specific and decision-relevant signals that SeqIA does not target, while its schema-guided design remains adaptable to different spatial granularity required for other hazards.

These insights point toward several practical recommendations for deploying such systems in real-world pipelines:

\begin{itemize}
    \item \textbf{Response parsing}: Enable only if models produce frequent formatting errors, and skip in newer LLM generations that emit valid JSON consistently. The role of response parsing will decline as model reliability improves.
    \item \textbf{Model size and precision}: Choose accordingly to latency and hardware budgets. Use quantized small/mid-size models for routine large-scale deployments, full precision only for critical use.
    \item \textbf{Model recency}: Favor latest LLM checkpoints, which consistently improve stability and accuracy.
    \item \textbf{Prompt design}: Adapt to capacity using shorter prompts for small models, and rich descriptions and additional guidance for larger ones.
    \item \textbf{Operational evaluation}: Where feasible, run a small-scale test with labeled data across candidate models to estimate the runtime–accuracy balance before scaling up.
    \item \textbf{Pipeline integration}: Combine supervised classifiers or alternative methods for fast relevance filtering with GenIE approaches for schema-flexible information extraction tasks like impacts and locations.
    \item \textbf{Geo-resolution}: Complement LLM predictions with deterministic geo-resolution methods (gazetteers, basin mapping, or province normalization) to mitigate recall issues and enforce spatial consistency.
    \item \textbf{Sustainability}: Prefer open-weight, small and quantized models whenever possible to minimize compute, memory, and energy footprints in line with sustainability principles~\cite{schwartzGreenAI2019,strubellEnergyPolicyConsiderations2019}.
\end{itemize}

Because CienaLLM is zero-shot rather than label-trained, its generalization prospects are strong. Extending from drought to floods, hail, or heatwaves requires only defining category descriptions, not relabeling corpora and retraining. The same flexibility applies across languages: prompts and schemas can be localized without retraining, enabling rapid adaptation to multilingual or regional outlets. Moderate domain shifts, such as evolving reporting styles or new terminology, can be addressed with lightweight monitoring and occasional audits, rather than expensive re-annotation campaigns. Our methodology's design decouples extraction logic from task-specific definition, making the framework portable across hazards, geographies, and languages.

To support reproducibility, we provide the full codebase, including the framework source code, configurations, and automation scripts. Although the datasets used in this study are not yet public, they will be released with a future peer-reviewed paper. Regarding validity, several steps were taken to ensure consistency: temperatures were fixed to zero and seeds set in the LLM backend to minimize stochasticity, and outputs that failed to parse were excluded to avoid formatting issues with extraction quality. Our evaluation is also centered on Spanish media, and transferability to other languages, outlets, or time periods may shift model priors and should therefore be assessed.

The study has some limitations. The experiments focused exclusively on droughts, leaving other hazards for future evaluation. Comparisons with SeqIA were constrained to the E2E dataset, given training overlaps. Statistical testing across configurations was limited, and more rigorous analyses could strengthen interpretation of performance differences. Finally, infrastructure benchmarks covered only a subset of consumer and HPC environments; a broader hardware study would provide a more complete picture of deployment costs.

\section{Conclusions} \label{sec:conclusions}

Extreme weather events, and droughts in particular, pose complex challenges that cannot be fully captured by physical indices alone. Understanding and monitoring their socio-economic consequences requires systematic approaches that integrate heterogeneous reportage into structured, decision-relevant signals.

In this work we introduced CienaLLM, a modular framework for schema-guided Generative Information Extraction (GenIE) using open-weight LLMs. Through a large-scale factorial evaluation spanning 384 configurations of model families, sizes, quantization regimes, and prompting strategies, we showed that CienaLLM can reliably extract drought impacts, classify relevance, and identify affected locations from Spanish news. Compared to the supervised SeqIA baseline~\cite{lopez-otalSeqIAPythonFramework2025}, CienaLLM delivers competitive or superior accuracy in several tasks, while offering unique advantages in schema-flexible extraction. Our analysis highlights key determinants of performance: larger models provide the most accurate and stable results, quantization delivers large efficiency gains with modest trade-offs, and prompt strategies act as tunable hyperparameters whose benefits depend on model family and scale. Location extraction remains the most difficult task, underscoring the need for further advances in handling implicit geographic references.

Beyond task-level findings, the broader significance of CienaLLM lies in its generalizability. Because the system is schema-guided rather than label-trained, extending the framework to new hazards (e.g., floods, hail, heatwaves), new impacts, or new geographic domains requires only adapting prompts and schema definitions, not relabeling corpora or retraining models. More generally, the same method can be applied outside the climate domain to extract other forms of structured information from heterogeneous text, underscoring the portability of the approach. Our design enables near–real-time generation of indicators from national-scale news. In doing so, CienaLLM bridges physical drought indices with the socio-economic consequences documented in media and provides decision-makers with timely evidence for adaptation planning.

Looking ahead, several directions remain open: extending evaluation to additional hazards and languages, developing ensemble strategies that combine complementary model families and sizes, and experimenting with alternative prompting paradigms such as few-shot or retrieval-augmented generation that might improve performance. Future work should also emphasize turning extracted outputs into actionable insights, tracing the evolution of drought impacts across sectors and regions.

Taken together, our results show that schema-guided GenIE with open-weight LLMs is a practical way to turn unstructured news into structured, decision-ready evidence of climate impacts. CienaLLM complements supervised pipelines rather than replacing them, offering schema flexibility, transparent configurations, and reproducible evaluation. Because the framework is portable across models, hazards, and languages, improvements in LLMs translate directly into better extraction without retraining. By releasing code and documenting the full experimental setup, we aim to make these gains cumulative and comparable. In short, CienaLLM helps move from scattered reportage to consistent indicators, a necessary step toward adaptive, data-driven climate-risk monitoring.

\section{Acknowledgments}

This work has been supported by the research projects TED2021-129152B-C42 and PID2022-137244OB-I00, financed by the Spanish Ministry of Science and FEDER, and CSIC's Interdisciplinary Thematic Platform Clima (PTI-Clima).

This research project was made possible through the access granted by the Galician Supercomputing Center (CESGA) to its supercomputing infrastructure. The supercomputer FinisTerrae III and its permanent data storage system have been funded by the NextGeneration EU 2021 Recovery, Transformation and Resilience Plan, ICT2021-006904, and also from the Pluriregional Operational Programme of Spain 2014-2020 of the European Regional Development Fund (ERDF), ICTS-2019-02-CESGA-3, and from the State Programme for the Promotion of Scientific and Technical Research of Excellence of the State Plan for Scientific and Technical Research and Innovation 2013-2016 State subprogramme for scientific and technical infrastructures and equipment of ERDF, CESG15-DE-3114.

\bibliographystyle{IEEEtran}
\bibliography{references}

\clearpage

\appendices

\makeatletter
\@addtoreset{figure}{section}
\@addtoreset{table}{section}
\renewcommand{\thefigure}{\thesection.\arabic{figure}}
\renewcommand{\thetable}{\thesection.\arabic{table}}
\makeatother

\section{Prompt Templates and Output Schema for Drought Impact Extraction} \label{sec:appendix_prompt_impacts}

This appendix contains the complete descriptions and exact templates used in the drought impact extraction task (see Section \ref{sec:experiments_impacts}). It documents the base prompt, all flag-specific modifications as well as the format instructions and the output schema. All prompts and schemas are assembled dynamically by the CienaLLM framework according to the selected configuration.

\begin{itemize}
    \item \textbf{Base Prompt}: Always included. Introduces the extraction task and inserts the article headline and body text (see Figure \ref{fig:appendix_prompt_impacts_base_prompt})
    \item \textbf{Impact Descriptions}: When enabling \texttt{DESC}, natural language definitions of each impact category are inserted into the base prompt to guide interpretation (see Figures \ref{fig:appendix_prompt_impacts_base_prompt_descriptions}, \ref{fig:appendix_prompt_impacts_agriculture}, \ref{fig:appendix_prompt_impacts_livestock}, \ref{fig:appendix_prompt_impacts_hydrological_resources}, and \ref{fig:appendix_prompt_impacts_energy}).
    \item \textbf{Zero-shot Chain-of-Thought}: When enabling \texttt{CoT}, appends an instruction to reason step-by-step before answering (see Figure \ref{fig:appendix_prompt_impacts_cot}).
    \item \textbf{Article Summarization}: When enabling \texttt{SUM}, performs an initial LLM call to summarize the article. The summary replaces the original article text in the main extraction prompt (see Figure \ref{fig:appendix_prompt_impacts_sum}).
    \item \textbf{Self-Criticism}: When enabling \texttt{SC}, the model's first output is passed to a second LLM call that reviews and corrects it if necessary (see Figure \ref{fig:appendix_prompt_impacts_sc}).
    \item \textbf{Response Parsing Strategy}: When enabling \texttt{RPARSE}, the initial model output is unstructured text. A second LLM call reformats it into structured JSON using the same impact descriptions and format instructions (see Figure \ref{fig:appendix_prompt_impacts_rparse}).
    \item \textbf{Format Instructions}: The JSON format instructions (see Figure \ref{fig:appendix_prompt_impacts_format}), derived from the output schema (see Figure \ref{fig:appendix_prompt_impacts_schema}), are inserted either directly in the main extraction prompt (\texttt{RPARSE = false}) or in the second parsing prompt (\texttt{RPARSE = true}).
\end{itemize}

\begin{figure}[htbp]
    \centering
    \begin{minipage}{0.8\textwidth}
        \begin{verbatim}
You are an expert in environmental
analysis.  Your task is to analyze the
following news article and determine
whether it reports or  mentions any
impact caused by {event} on specific
aspects.

The aspects to consider are: {impacts}

Please carefully read the article and
determine for each aspect whether there
is a reported impact caused specifically
by {event}. Do not infer impacts unless
they are clearly stated or strongly
implied in the text.

Article to analyze:
{text}
        \end{verbatim}
    \end{minipage}
    \caption{Base Prompt for Drought Impact Extraction.}
    \label{fig:appendix_prompt_impacts_base_prompt}
\end{figure}

\begin{figure}[htbp]
    \centering
    \begin{minipage}{0.8\textwidth}
        \begin{verbatim}
You are an expert in environmental
analysis. Your task is to analyze the
following news article and determine
whether it reports or mentions any
impact caused by {event} on specific
aspects.

The aspects to consider are: {impacts}

Each aspect is briefly described below to
guide interpretation, but these
definitions are not exhaustive:

{impact_descriptions}

Please carefully read the article and
determine for each aspect whether there
is a reported impact caused specifically
by {event}. Do not infer impacts unless
they are clearly stated or strongly
implied in the text.

Article to analyze:
{text}
       \end{verbatim}
    \end{minipage}
    \caption{Base Prompt with Descriptions for Drought Impact Extraction.}
    \label{fig:appendix_prompt_impacts_base_prompt_descriptions}
\end{figure}

\begin{figure}[htbp]
    \centering
    \begin{minipage}{0.8\textwidth}
        \begin{verbatim}
News about the impacts of drought on
agriculture usually refer to losses in
both rainfed and irrigated crops. It is
often mentioned that part of the harvest
has been lost or will be lost.
       \end{verbatim}
    \end{minipage}
    \caption{Description of Drought Impact ``Agriculture''.}
    \label{fig:appendix_prompt_impacts_agriculture}
\end{figure}

\begin{figure}[htbp]
    \centering
    \begin{minipage}{0.8\textwidth}
        \begin{verbatim}
News about the impacts of drought on
livestock usually refer to the loss of
pastures that feed the livestock. In more
extreme droughts, it may be mentioned that
there is no water available to give the
livestock to drink.
       \end{verbatim}
    \end{minipage}
    \caption{Description of Drought Impact ``Livestock''.}
    \label{fig:appendix_prompt_impacts_livestock}
\end{figure}

\begin{figure}[htbp]
    \centering
    \begin{minipage}{0.8\textwidth}
        \begin{verbatim}
News about the impacts of drought on
hydrological resources usually mention
the decrease in river flows, reservoir
levels, or groundwater. It is also
mentioned the lack of water that this
causes in the populations, with water
cuts being decreed or water not being
used for certain uses, or the need to
bring water from other locations.
       \end{verbatim}
    \end{minipage}
    \caption{Description of Drought Impact ``Hydrological Resources''.}
    \label{fig:appendix_prompt_impacts_hydrological_resources}
\end{figure}

\begin{figure}[htbp]
    \centering
    \begin{minipage}{0.8\textwidth}
        \begin{verbatim}
News about the impacts of drought on
energy usually mention that due to the
low flow of rivers or reservoirs, it is
not possible to turbine and therefore
the generation of hydroelectric energy
decreases.
       \end{verbatim}
    \end{minipage}
    \caption{Description of Drought Impact ``Energy''.}
    \label{fig:appendix_prompt_impacts_energy}
\end{figure}

\begin{figure}[htbp]
    \centering
    \begin{minipage}{0.8\textwidth}
        \begin{verbatim}
Reason step by step and explain your
reasoning before giving the final answer.
       \end{verbatim}
    \end{minipage}
    \caption{Chain-of-Thought Instruction.}
    \label{fig:appendix_prompt_impacts_cot}
\end{figure}

\begin{figure}[htbp]
    \centering
    \begin{minipage}{0.8\textwidth}
        \begin{verbatim}
Summarize the following article.

Text:
{text}
       \end{verbatim}
    \end{minipage}
    \caption{Summarization Prompt.}
    \label{fig:appendix_prompt_impacts_sum}
\end{figure}

\begin{figure}[htbp]
    \centering
    \begin{minipage}{0.8\textwidth}
        \begin{verbatim}
Given the following prompt:
{prompt}

And the following response:
{response}

Analyze the response and determine
whether it is correct or incorrect. If
it is incorrect, provide a brief
explanation of why it is incorrect and
the correct response.
If it is correct, provide the same correct
response.
       \end{verbatim}
    \end{minipage}
    \caption{Self-Criticism Prompt.}
    \label{fig:appendix_prompt_impacts_sc}
\end{figure}

\begin{figure}[htbp]
    \centering
    \begin{minipage}{0.8\textwidth}
        \begin{verbatim}
Extract whether the following LLM
response says the article mentions an
impact of {event} on {impacts}.

The impacts are defined as follows:
{impact_descriptions}

Text:
{text}
       \end{verbatim}
    \end{minipage}
    \caption{Response Parsing Prompt for Drought Impact Extraction.}
    \label{fig:appendix_prompt_impacts_rparse}
\end{figure}

\begin{figure}[htbp]
    \centering
    \begin{minipage}{0.8\textwidth}
        \begin{verbatim}
Format instructions:
{format_instructions}
Make sure to include a single JSON in your
response instead of multiple JSONs.
       \end{verbatim}
    \end{minipage}
    \caption{Response Parsing Format Instruction.}
    \label{fig:appendix_prompt_impacts_format}
\end{figure}

\begin{figure}[htbp]
    \centering
    \begin{minipage}{0.8\textwidth}
        \begin{verbatim}
{
    "agriculture": <true or false>,
    "livestock": <true or false>,
    "hydrological_resources":
        <true or false>,
    "energy": <true or false>
}
       \end{verbatim}
    \end{minipage}
    \caption{Output Schema for Drought Impact Extraction.}
    \label{fig:appendix_prompt_impacts_schema}
\end{figure}

\section{Prompt Templates and Output Schema for Drought Relevance Classification} \label{sec:appendix_prompt_relevance}

This appendix describes the prompt template and output schema used for the \textbf{drought relevance classification} task (see Section \ref{sec:experiments_relevance}). The setup follows the same modular structure, format instructions, and optional response parsing mechanisms described in Appendix \ref{sec:appendix_prompt_impacts}. The base prompt asks whether the article mentions a drought-related event, as shown in Figure \ref{fig:appendix_prompt_relevance_base}. When \texttt{RPARSE = true}, a second prompt is used to reformat the initial output into structured JSON, illustrated in Figure \ref{fig:appendix_prompt_relevance_rparse}. The final output is expected to follow a simplified schema consisting of a single boolean field \texttt{"drought"}, shown in Figure \ref{fig:appendix_prompt_relevance_schema}.

\begin{figure}[htbp]
    \centering
    \begin{minipage}{0.8\textwidth}
        \begin{verbatim}
Analyze the following article and
determine if the news article mentions
an event related to {event}.

Text:
{text}
       \end{verbatim}
    \end{minipage}
    \caption{Base Prompt for Drought Relevance Classification.}
    \label{fig:appendix_prompt_relevance_base}
\end{figure}

\begin{figure}[htbp]
    \centering
    \begin{minipage}{0.8\textwidth}
        \begin{verbatim}
Extract whether the following LLM
response says the article mentions an
event related to {event}.

Text:
{text}
       \end{verbatim}
    \end{minipage}
    \caption{Response Parsing Prompt for Drought Relevance Classification.}
    \label{fig:appendix_prompt_relevance_rparse}
\end{figure}

\begin{figure}[htbp]
    \centering
    \begin{minipage}{0.8\textwidth}
        \begin{verbatim}
{
    "drought": <true or false>
}
       \end{verbatim}
    \end{minipage}
    \caption{Output Schema for Drought Relevance Classification.}
    \label{fig:appendix_prompt_relevance_schema}
\end{figure}

\section{Prompt Templates and Output Schema for Drought Impact Location Extraction} \label{sec:appendix_prompt_locations}

This appendix describes the prompt template and output schema used for the drought impact location extraction task (see Section \ref{sec:experiments_locations}). It builds on the same modular construction and response parsing logic described in Appendix \ref{sec:appendix_prompt_impacts}, with task-specific modifications. The base prompt instructs the model to identify Spanish provinces affected by drought, as shown in Figure \ref{fig:appendix_prompt_locations_base}. If response parsing is enabled, a second prompt is used to convert the output into a valid JSON structure, as seen in Figure \ref{fig:appendix_prompt_locations_rparse}. The expected schema consists of a field "provinces" containing a list of province names, illustrated in Figure \ref{fig:appendix_prompt_locations_schema}.

\begin{figure}[htbp]
    \centering
    \begin{minipage}{0.8\textwidth}
        \begin{verbatim}
Given a news article describing {event}
impacts in Spanish regions, return the
list of affected provinces.
Identify provinces explicitly mentioned
or infer them from the described
locations.
If you cannot identify specific
provinces, return all provinces in the
regions mentioned.
Note that the text may refer to
autonomous communities or specific
cities, towns, and municipalities,
which are not the provinces being
requested. Do not include these
autonomous communities or
municipalities in the output.
Return the province names in Spanish.

Text:
{text}
       \end{verbatim}
    \end{minipage}
    \caption{Base Prompt for Drought Impact Location Extraction}
    \label{fig:appendix_prompt_locations_base}
\end{figure}

\begin{figure}[htbp]
    \centering
    \begin{minipage}{0.8\textwidth}
        \begin{verbatim}
Extract from the following LLM response
the provinces.
If other locations are mentioned, infer
the provinces.

Text:
{text}
       \end{verbatim}
    \end{minipage}
    \caption{Response Parsing Prompt for Drought Impact Location Extraction}
    \label{fig:appendix_prompt_locations_rparse}
\end{figure}

\begin{figure}[htbp]
    \centering
    \begin{minipage}{0.8\textwidth}
        \begin{verbatim}
{
    "response": [
        <province>,
        ...
    ]
}
       \end{verbatim}
    \end{minipage}
    \caption{Output Schema for Drought Impact Location Extraction}
    \label{fig:appendix_prompt_locations_schema}
\end{figure}

\section{Results Tables} \label{sec:appendix_results}

This appendix compiles the tables referenced in Section~\ref{sec:results}.
Table~\ref{tab:appendix_results_rparse} reports the overall comparison of average performance with and without response parsing (\texttt{RPARSE}), while Table~\ref{tab:model_metrics_response_parsing} presents the corresponding model-level metrics.
Table~\ref{tab:appendix_results_model} summarizes average performance and variability per LLM across all prompt configurations, and Table~\ref{tab:appendix_results_model_top} lists the best-performing configuration for each model.
Table~\ref{tab:appendix_results_quantization} compares full-precision and quantized variants across model families.
Table~\ref{tab:appendix_results_prompt} shows average performance by configuration factor, with Tables~\ref{tab:appendix_results_prompt_model} and~\ref{tab:appendix_results_prompt_family} detailing factor effects at the model and family levels, respectively.
Finally, Table~\ref{tab:appendix_results_tradeoff} presents the three Pareto-optimal configurations used throughout the efficiency–performance analysis.

For discussion and interpretation of these results, see Section~\ref{sec:results}.

\begin{table*}[!t]
    \centering
    \caption{Average drought impact extraction performance with and without \texttt{RPARSE}.}
    \label{tab:appendix_results_rparse}
    \small
    \begin{tabular}{|c|S[table-format=1.3(3)]|S[table-format=1.3(3)]|S[table-format=1.3(3)]|}
        \hline
        \textbf{\texttt{RPARSE}}      & {\textbf{F1 Score}}                     & {\textbf{Parsing Error Rate}}                                  & {\textbf{Exec. Time (s/article)}}                             \\
        \hline
        \texttt{False}                & 0.803(61)                               & 0.040(83)                                                      & 10.175(10575)                                                 \\
        \texttt{True}                 & 0.808(48)                               & 0.007(17)                                                      & 16.096(15044)                                                 \\
        \hline
        \textbf{\(\Delta\) (p-value)} & \multicolumn{1}{c|}{+0.005 ($p = 1.0$)} & \multicolumn{1}{c|}{--0.033$^{***}$ ($p = 5.5\times10^{-13}$)} & \multicolumn{1}{c|}{+5.921$^{***}$ ($p = 9.7\times10^{-32}$)} \\
        \hline
    \end{tabular}

    \vspace{2mm}
    \raggedright \footnotesize \textit{Note.} Significance stars are used consistently across all tables: $^*$\,$p < 0.05$; $^{**}$\,$p < 0.01$; $^{***}$\,$p < 0.001$.

\end{table*}

\begin{table*}[!t]
    \centering
    \caption{Average drought impact extraction performance per model with and without \texttt{RPARSE}.}
    \label{tab:model_metrics_response_parsing}
    \small
    \begin{tabular}{|l|c|S[table-format=1.3(3)]|S[table-format=1.3(3)]|S[table-format=1.3(3)]|}
        \hline
        \textbf{\texttt{LLM}}  & \textbf{\texttt{RPARSE}} & {\textbf{F1 Score}} & {\textbf{Parsing Error Rate}} & {\textbf{Exec. Time (s/article)}} \\
        \hline
        \texttt{gemma\_4b}     & \texttt{False}           & 0.750(56)           & 0.014(17)                     & 4.461(200)                        \\
                               & \texttt{True}            & 0.727(57)           & 0.000(0)                      & 6.875(146)                        \\
        \texttt{gemma\_12b}    & \texttt{False}           & 0.796(32)           & 0.003(4)                      & 6.908(329)                        \\
                               & \texttt{True}            & 0.811(30)           & 0.000(0)                      & 12.902(386)                       \\
        \texttt{gemma\_12b\_f} & \texttt{False}           & 0.805(28)           & 0.000(0)                      & 10.181(516)                       \\
                               & \texttt{True}            & 0.818(28)           & 0.000(0)                      & 17.688(519)                       \\
        \texttt{gemma\_27b}    & \texttt{False}           & 0.838(23)           & 0.010(24)                     & 11.411(552)                       \\
                               & \texttt{True}            & 0.833(22)           & 0.000(0)                      & 18.272(552)                       \\
        \texttt{llama\_3b}     & \texttt{False}           & 0.776(51)           & 0.135(139)                    & 3.553(132)                        \\
                               & \texttt{True}            & 0.766(25)           & 0.000(0)                      & 4.939(148)                        \\
        \texttt{llama\_8b}     & \texttt{False}           & 0.793(53)           & 0.186(123)                    & 5.312(202)                        \\
                               & \texttt{True}            & 0.813(17)           & 0.044(8)                      & 7.054(201)                        \\
        \texttt{llama\_8b\_f}  & \texttt{False}           & 0.758(146)          & 0.104(55)                     & 8.415(335)                        \\
                               & \texttt{True}            & 0.830(15)           & 0.041(19)                     & 10.762(314)                       \\
        \texttt{llama\_70b}    & \texttt{False}           & 0.855(14)           & 0.021(33)                     & 27.754(16993)                     \\
                               & \texttt{True}            & 0.858(9)            & 0.000(0)                      & 41.631(13327)                     \\
        \texttt{qwen\_3b}      & \texttt{False}           & 0.773(27)           & 0.007(12)                     & 3.731(208)                        \\
                               & \texttt{True}            & 0.747(39)           & 0.000(0)                      & 6.202(229)                        \\
        \texttt{qwen\_7b}      & \texttt{False}           & 0.819(20)           & 0.000(1)                      & 5.165(199)                        \\
                               & \texttt{True}            & 0.815(30)           & 0.000(0)                      & 7.134(247)                        \\
        \texttt{qwen\_7b\_f}   & \texttt{False}           & 0.826(9)            & 0.002(3)                      & 8.280(342)                        \\
                               & \texttt{True}            & 0.823(25)           & 0.000(0)                      & 11.971(458)                       \\
        \texttt{qwen\_72b}     & \texttt{False}           & 0.846(15)           & 0.000(0)                      & 26.928(14112)                     \\
                               & \texttt{True}            & 0.852(15)           & 0.000(0)                      & 47.725(15842)                     \\
        \hline
    \end{tabular}
\end{table*}

\begin{table*}[!t]
    \centering
    \caption{Average drought impact extraction performance per model.}
    \label{tab:appendix_results_model}
    \small
    \begin{tabular}{|l|S[table-format=1.3(3)]|S[table-format=1.3(3)]|S[table-format=1.3(3)]|}
        \hline
        \textbf{\texttt{LLM}}  & {\textbf{F1 Score}} & {\textbf{Parsing Error Rate}} & {\textbf{Exec. Time (s/article)}} \\
        \hline
        \texttt{gemma\_4b}     & 0.727(57)           & 0.000(0)                      & 6.875(146)                        \\
        \texttt{gemma\_12b}    & 0.811(30)           & 0.000(0)                      & 12.902(386)                       \\
        \texttt{gemma\_12b\_f} & 0.818(28)           & 0.000(0)                      & 17.688(519)                       \\
        \texttt{gemma\_27b}    & 0.833(22)           & 0.000(0)                      & 18.272(552)                       \\
        \texttt{llama\_3b}     & 0.766(25)           & 0.000(0)                      & 4.939(148)                        \\
        \texttt{llama\_8b}     & 0.813(17)           & 0.044(8)                      & 7.054(201)                        \\
        \texttt{llama\_8b\_f}  & 0.830(15)           & 0.041(19)                     & 10.762(314)                       \\
        \texttt{llama\_70b}    & 0.858(9)            & 0.000(0)                      & 41.631(1333)                      \\
        \texttt{qwen\_3b}      & 0.747(39)           & 0.000(0)                      & 6.202(229)                        \\
        \texttt{qwen\_7b}      & 0.815(30)           & 0.000(0)                      & 7.134(247)                        \\
        \texttt{qwen\_7b\_f}   & 0.823(25)           & 0.000(0)                      & 11.971(458)                       \\
        \texttt{qwen\_72b}     & 0.852(15)           & 0.000(0)                      & 47.725(1584)                      \\
        \hline
    \end{tabular}
\end{table*}

\begin{table*}[!t]
    \centering
    \caption{Drought impact extraction performance for best-performing configuration per model.}
    \label{tab:appendix_results_model_top}
    \small
    \begin{tabular}{|l|l|S[table-format=1.3]|S[table-format=1.3]|S[table-format=1.3]|}
        \hline
        \textbf{\texttt{LLM}}  & \textbf{Prompt Techniques}                  & {\textbf{F1 Score}} & {\textbf{Parsing Error Rate}} & {\textbf{Exec. Time (s/article)}} \\
        \hline
        \texttt{gemma\_4b}     & \texttt{SUM} + \texttt{CoT} + \texttt{DESC} & 0.796               & 0.000                         & 6.591                             \\
        \texttt{gemma\_12b}    & \texttt{SUM} + \texttt{DESC}                & 0.851               & 0.000                         & 10.514                            \\
        \texttt{gemma\_12b\_f} & \texttt{SUM} + \texttt{CoT} + \texttt{DESC} & 0.855               & 0.000                         & 16.169                            \\
        \texttt{gemma\_27b}    & \texttt{SUM} + \texttt{CoT} + \texttt{DESC} & 0.865               & 0.000                         & 16.816                            \\
        \texttt{llama\_3b}     & \texttt{-}                                  & 0.798               & 0.000                         & 2.724                             \\
        \texttt{llama\_8b}     & \texttt{CoT} + \texttt{DESC}                & 0.838               & 0.048                         & 5.388                             \\
        \texttt{llama\_8b\_f}  & \texttt{DESC}                               & 0.856               & 0.026                         & 5.321                             \\
        \texttt{llama\_70b}    & \texttt{DESC}                               & 0.871               & 0.000                         & 21.974                            \\
        \texttt{qwen\_3b}      & \texttt{-}                                  & 0.797               & 0.000                         & 2.741                             \\
        \texttt{qwen\_7b}      & \texttt{CoT}                                & 0.849               & 0.000                         & 5.257                             \\
        \texttt{qwen\_7b\_f}   & \texttt{CoT} + \texttt{DESC}                & 0.846               & 0.000                         & 8.589                             \\
        \texttt{qwen\_72b}     & \texttt{CoT} + \texttt{DESC}                & 0.878               & 0.000                         & 35.793                            \\
        \hline
    \end{tabular}
\end{table*}

\begin{table*}[!t]
    \centering
    \caption{Average drought impact extraction performance per model family with quantized and full-precision.}
    \label{tab:appendix_results_quantization}
    \small
    \begin{tabular}{|l|c|S[table-format=1.3(3)]|S[table-format=1.3(3)]|S[table-format=2.3(3)]|}
        \hline
        \textbf{\texttt{LLM} Family}    & \textbf{\texttt{LLM} Quantization} & {\textbf{F1 Score}}                               & {\textbf{Parsing Error Rate}}            & {\textbf{Exec. Time (s/article)}}                    \\
        \hline
        \multirow{3}{*}{\texttt{gemma}} & \texttt{fp16}                      & 0.818(27)                                         & 0.000(0)                                 & 17.688(5028)                                         \\
                                        & \texttt{q4\_K\_M}                  & 0.811(29)                                         & 0.000(0)                                 & 12.902(3741)                                         \\
                                        & \textbf{\(\Delta\) (p-value)}      & \multicolumn{1}{c|}{--0.008$^{**}$ ($p = 0.005$)} & \multicolumn{1}{c|}{0.000 ($p =$ ---)}   & \multicolumn{1}{c|}{--4.786$^{***}$ ($p < 10^{-4}$)} \\
        \hline
        \multirow{3}{*}{\texttt{llama}} & \texttt{fp16}                      & 0.830(15)                                         & 0.041(19)                                & 10.762(3040)                                         \\
                                        & \texttt{q4\_K\_M}                  & 0.813(16)                                         & 0.044(8)                                 & 7.054(1945)                                          \\
                                        & \textbf{\(\Delta\) (p-value)}      & \multicolumn{1}{c|}{--0.017$^{**}$ ($p = 0.005$)} & \multicolumn{1}{c|}{0.003 ($p = 0.469$)} & \multicolumn{1}{c|}{--3.708$^{***}$ ($p < 10^{-4}$)} \\
        \hline
        \multirow{3}{*}{\texttt{qwen}}  & \texttt{fp16}                      & 0.823(25)                                         & 0.000(0)                                 & 11.971(4434)                                         \\
                                        & \texttt{q4\_K\_M}                  & 0.815(29)                                         & 0.000(0)                                 & 7.134(2393)                                          \\
                                        & \textbf{\(\Delta\) (p-value)}      & \multicolumn{1}{c|}{--0.008$^{*}$ ($p = 0.010$)}  & \multicolumn{1}{c|}{0.000 ($p =$ ---)}   & \multicolumn{1}{c|}{--4.837$^{***}$ ($p < 10^{-4}$)} \\
        \hline
    \end{tabular}
    \vspace{1mm}
\end{table*}

\begin{table*}[!t]
    \centering
    \caption{Average drought impact extraction performance per prompt strategy.}
    \label{tab:appendix_results_prompt}
    \small
    \begin{tabular}{|l|c|S[table-format=1.3(3)]|S[table-format=1.3(3)]|S[table-format=1.3(3)]|}
        \hline
        \textbf{Configuration}         & \textbf{Value}                & {\textbf{F1 Score}}                      & {\textbf{Parsing Error Rate}}                  & {\textbf{Exec. Time (s/article)}}                   \\
        \hline
        \multirow{3}{*}{\texttt{SUM}}  & \texttt{False}                & 0.807(55)                                & 0.007(17)                                      & 14.032(13418)                                       \\
                                       & \texttt{True}                 & 0.809(40)                                & 0.007(17)                                      & 18.161(16178)                                       \\
                                       & \textbf{\(\Delta\) (p-value)} & \multicolumn{1}{c|}{0.002 ($p = 1.0$)}   & \multicolumn{1}{c|}{0.000 ($p = 1.0$)}         & \multicolumn{1}{c|}{4.129$^{***}$ ($p < 10^{-16}$)} \\
        \hline
        \multirow{3}{*}{\texttt{SC}}   & \texttt{False}                & 0.808(48)                                & 0.007(17)                                      & 12.202(10803)                                       \\
                                       & \texttt{True}                 & 0.808(48)                                & 0.007(17)                                      & 19.990(17415)                                       \\
                                       & \textbf{\(\Delta\) (p-value)} & \multicolumn{1}{c|}{--0.000 ($p = 1.0$)} & \multicolumn{1}{c|}{0.000 ($p =$ ---)}         & \multicolumn{1}{c|}{7.788$^{***}$ ($p < 10^{-16}$)} \\
        \hline
        \multirow{3}{*}{\texttt{CoT}}  & \texttt{False}                & 0.806(48)                                & 0.006(13)                                      & 14.835(13575)                                       \\
                                       & \texttt{True}                 & 0.809(47)                                & 0.009(20)                                      & 17.358(16212)                                       \\
                                       & \textbf{\(\Delta\) (p-value)} & \multicolumn{1}{c|}{0.002 ($p = 0.186$)} & \multicolumn{1}{c|}{0.003$^{*}$ ($p = 0.014$)} & \multicolumn{1}{c|}{2.523$^{***}$ ($p < 10^{-8}$)}  \\
        \hline
        \multirow{3}{*}{\texttt{DESC}} & \texttt{False}                & 0.807(39)                                & 0.006(15)                                      & 16.178(14732)                                       \\
                                       & \texttt{True}                 & 0.809(55)                                & 0.008(19)                                      & 16.015(15272)                                       \\
                                       & \textbf{\(\Delta\) (p-value)} & \multicolumn{1}{c|}{0.002 ($p = 0.510$)} & \multicolumn{1}{c|}{0.002$^{*}$ ($p = 0.013$)} & \multicolumn{1}{c|}{--0.163 ($p = 1.0$)}            \\
        \hline
    \end{tabular}
    \vspace{1mm}
\end{table*}

\begin{table*}[!t]
    \centering
    \caption{Average effect size (\(\Delta\)) on drought impact extraction performance per model per prompt strategy.}
    \label{tab:appendix_results_prompt_model}
    \small
    \begin{tabular}{|l|
        S[table-format=+1.3]@{\hspace{1.5em}}l|
        S[table-format=+1.3]@{\hspace{1.5em}}l|
        S[table-format=+1.3]@{\hspace{1.5em}}l|
        S[table-format=+1.3]@{\hspace{1.5em}}l|}
        \hline
        \textbf{\texttt{LLM}}  & \multicolumn{2}{c|}{\textbf{\texttt{SUM}}} & \multicolumn{2}{c|}{\textbf{\texttt{SC}}} & \multicolumn{2}{c|}{\textbf{\texttt{CoT}}} & \multicolumn{2}{c|}{\textbf{\texttt{DESC}}}                                                                 \\
        \hline
        \texttt{gemma\_4b}     & +0.110$^{**}$                              & ($p = 0.008$)                             & -0.000                                     & ($p = 0.144$)                               & +0.006$^{*}$  & ($p = 0.023$) & +0.009        & ($p = 0.312$) \\
        \texttt{gemma\_12b}    & +0.040$^{**}$                              & ($p = 0.008$)                             & -0.002                                     & ($p = 0.249$)                               & -0.002        & ($p = 0.742$) & +0.039$^{**}$ & ($p = 0.008$) \\
        \texttt{gemma\_12b\_f} & +0.039$^{**}$                              & ($p = 0.008$)                             & +0.001                                     & ($p = 0.180$)                               & +0.006        & ($p = 0.148$) & +0.035$^{**}$ & ($p = 0.008$) \\
        \texttt{gemma\_27b}    & +0.028$^{**}$                              & ($p = 0.008$)                             & +0.000                                     & ($p = 0.180$)                               & +0.010$^{**}$ & ($p = 0.008$) & +0.029$^{**}$ & ($p = 0.008$) \\
        \texttt{llama\_3b}     & -0.014                                     & ($p = 0.148$)                             & +0.000                                     & ($p = $---)                                 & -0.013$^{**}$ & ($p = 0.008$) & -0.042$^{**}$ & ($p = 0.008$) \\
        \texttt{llama\_8b}     & -0.029$^{**}$                              & ($p = 0.008$)                             & +0.000                                     & ($p = $---)                                 & -0.005        & ($p = 0.312$) & +0.005        & ($p = 0.383$) \\
        \texttt{llama\_8b\_f}  & -0.011                                     & ($p = 0.148$)                             & +0.000                                     & ($p = $---)                                 & +0.004        & ($p = 0.312$) & +0.022$^{**}$ & ($p = 0.008$) \\
        \texttt{llama\_70b}    & -0.012$^{**}$                              & ($p = 0.008$)                             & +0.000                                     & ($p = $---)                                 & -0.001        & (p = 1.000)   & +0.009$^{*}$  & ($p = 0.039$) \\
        \texttt{qwen\_3b}      & -0.019$^{**}$                              & ($p = 0.008$)                             & +0.000                                     & ($p = $---)                                 & +0.002        & ($p = 0.742$) & -0.070$^{**}$ & ($p = 0.008$) \\
        \texttt{qwen\_7b}      & -0.053$^{**}$                              & ($p = 0.008$)                             & +0.000                                     & ($p = $---)                                 & +0.008$^{**}$ & ($p = 0.008$) & -0.019$^{**}$ & ($p = 0.008$) \\
        \texttt{qwen\_7b\_f}   & -0.043$^{**}$                              & ($p = 0.008$)                             & +0.000                                     & ($p = $---)                                 & +0.013$^{**}$ & ($p = 0.008$) & -0.011        & ($p = 0.148$) \\
        \texttt{qwen\_72b}     & -0.014$^{*}$                               & ($p = 0.039$)                             & +0.000                                     & ($p = $---)                                 & +0.001        & ($p = 0.742$) & +0.021$^{**}$ & ($p = 0.008$) \\
        \hline
    \end{tabular}
    \vspace{1mm}
\end{table*}

\begin{table*}[!t]
    \centering
    \caption{Average effect size (\(\Delta\)) on drought impact extraction performance per model family per prompt strategy.}
    \label{tab:appendix_results_prompt_family}
    \small
    \begin{tabular}{|l|
        S[table-format=+1.3]@{\hspace{1.5em}}l|
        S[table-format=+1.3]@{\hspace{1.5em}}l|
        S[table-format=+1.3]@{\hspace{1.5em}}l|
        S[table-format=+1.3]@{\hspace{1.5em}}l|}
        \hline
        \textbf{\texttt{LLM} Family} & \multicolumn{2}{c|}{\textbf{\texttt{SUM}}} & \multicolumn{2}{c|}{\textbf{\texttt{SC}}} & \multicolumn{2}{c|}{\textbf{\texttt{CoT}}} & \multicolumn{2}{c|}{\textbf{\texttt{DESC}}}                                                                  \\
        \hline
        \texttt{gemma}               & +0.054$^{***}$                             & ($p = 0.000$)                             & -0.000                                     & ($p = 0.826$)                               & +0.005$^{**}$ & ($p = 0.004$) & +0.028$^{***}$ & ($p = 0.000$) \\
        \texttt{llama}               & -0.016$^{***}$                             & ($p = 0.000$)                             & +0.000                                     & ($p = $---)                                 & -0.004        & ($p = 0.080$) & -0.001         & ($p = 0.747$) \\
        \texttt{qwen}                & -0.032$^{***}$                             & ($p = 0.000$)                             & +0.000                                     & ($p = $---)                                 & +0.006$^{*}$  & ($p = 0.019$) & -0.020$^{**}$  & (p = 0.007)   \\
        \hline
    \end{tabular}
    \vspace{1mm}
\end{table*}

\begin{table*}[!t]
    \centering
    \caption{Drought impact extraction performance for top configurations on the Pareto front.}
    \label{tab:appendix_results_tradeoff}
    \small
    \begin{tabular}{|l|l|l|
            S[table-format=1.3]|
            S[table-format=1.3]|
            S[table-format=2.2]|}
        \hline
        \textbf{Configuration} & \textbf{\texttt{LLM}} & \textbf{Prompt Techniques}   & \textbf{F1 Score} & \textbf{Parse Error Rate} & \textbf{Exec. Time (s/article)} \\
        \hline
        \textit{Best-F1}       & \texttt{qwen\_72b}    & \texttt{CoT} + \texttt{DESC} & 0.878             & 0.000                     & 35.792                          \\
        \textit{FASTEST}       & \texttt{qwen\_3b}     & \texttt{DESC}                & 0.726             & 0.000                     & 2.633                           \\
        \textit{Efficient}     & \texttt{qwen\_7b}     & ---                          & 0.844             & 0.000                     & 3.243                           \\
        \hline
    \end{tabular}
\end{table*}

\section{News Corpora} \label{sec:appendix_news}

We collected broad news corpora by crawling the full online archives of three major Spanish outlets: El País, ABC, and 20~Minutos. We summarize their coverage in Table~\ref{tab:appendix_news}. In addition, we incorporate a previously compiled collection of drought-related articles from Grupo Zeta (a media conglomerate now integrated into Prensa Ibérica), originally assembled by López-Otal et~al.~\cite{lopez-otalSeqIAPythonFramework2025} using a keyword-based search focused on drought terms~\cite{lopez-otalSeqIAPythonFramework2025}. Unlike our outlet-wide crawls (which include \emph{all} articles in the specified ranges), the Grupo Zeta collection is \emph{task-focused} (drought-only) and not exhaustive over the full archive period reported in Table~\ref{tab:appendix_news}.

To reflect distinct access modalities at ABC, we report two entries: (i) the HTML archive crawl and (ii) the digitized historical \emph{hemeroteca} (PDF) crawl. Together with El País and 20~Minutos, these national outlets provide long temporal coverage suitable for retrospective analyses. Crucially, the scale and breadth of these corpora make them well suited for large-scale, longitudinal assessments of climate-related impacts in Spain using our CienaLLM extraction framework.

\begin{table*}[!t]
    \caption{Overview of news articles corpora.}
    \label{tab:appendix_news}
    \centering
    \begin{tabular}{|l|l|l|l|r|}
        \hline
        \textbf{News Outlet} & \textbf{Extraction Method}            & \textbf{Date Range}      & \textbf{URL}                           & \textbf{\# Articles} \\
        \hline
        El País              & All news articles (HTML archive)      & 1976-05-03 -- 2024-08-19 & \url{https://elpais.com/archivo/}      & 3{,}733{,}758        \\
        ABC                  & All news articles (HTML archive)      & 2001-01-10 -- 2024-11-01 & \url{https://www.abc.es/archivo/}      & 3{,}826{,}855        \\
        ABC (PDF)            & All news articles (PDF)               & 1891-05-10 -- 2024-10-18 & \url{https://www.abc.es/hemeroteca/}   & 6{,}435{,}816        \\
        20~Minutos           & All news articles (HTML archive)      & 2005-01-16 -- 2024-12-31 & \url{https://www.20minutos.es/archivo} & 2{,}631{,}446        \\
        Grupo Zeta           & Drought keyword search (task-focused) & 2004-02-06 -- 2022-11-28 & \textit{N/A}                           & 52{,}495             \\
        \hline
    \end{tabular}
\end{table*}

\end{document}